\documentclass[10pt,journal,final]{IEEEtran}
\usepackage{times}
\usepackage{latexsym}
\usepackage{multirow}
\usepackage{graphicx}
\usepackage{makecell}
\usepackage{amssymb}
\usepackage{amsmath}
\usepackage{xcolor}
\usepackage{array}
\usepackage{pifont}
\usepackage{booktabs}

\usepackage{arydshln}
\usepackage[ruled,linesnumbered]{algorithm2e}
\newcommand{\zqh}[1]{{\color{black} #1}}

\begin{document}
%
\title{Unified Instance and Knowledge Alignment Pretraining for Aspect-based Sentiment Analysis}
%
%
%

\author{Juhua~Liu,~\IEEEmembership{Member,~IEEE,}
        Qihuang~Zhong,~\IEEEmembership{Member,~IEEE,}
        Liang~Ding,~\IEEEmembership{Member,~IEEE,}
        Hua Jin,
        Bo~Du,~\IEEEmembership{Senior~Member,~IEEE,}
         and~Dacheng~Tao,~\IEEEmembership{Fellow,~IEEE}
\thanks{This work was supported in part by National Key Research and Development Program of China under Grant 2021YFB2401302, in part by the National Natural Science Foundation of China under Grant 62076186 and 62225113, and in part by Science and Technology Major Project of Hubei Province (Next-Generation AI Technologies) under Grant 2019AEA170. The numerical calculations in this paper have been done on the supercomputing system in the Supercomputing Center of Wuhan University. \textit{Juhua Liu and Qihuang Zhong contribute equally to this work. Corresponding Author: Bo Du (e-mail: dubo@whu.edu.cn).}}

\thanks{J. Liu is with the the Research Center for Graphic Communication, Printing and Packaging, and Institute of Artificial Intelligence, Wuhan University, Wuhan, China (e-mail: liujuhua@whu.edu.cn).}

\thanks{Q. Zhong and B. Du are  with the National Engineering Research Center for Multimedia Software, Institute of Artificial Intelligence, School of Computer Science and Hubei Key Laboratory of Multimedia and Network Communication Engineering, Wuhan University, Wuhan, China (e-mail: zhongqihuang@whu.edu.cn; dubo@whu.edu.cn).}

\thanks{L. Ding is with the JD Explore Academy at JD.com, Beijing, China (e-mail: liangding.liam@gmail.com).}

\thanks{H. Jin is with the affiliated hospital of Kunming University of Science and Technology, Kunming, China (e-mail: jinhuakm@163.com).}

\thanks{D. Tao is with the School of Computer Science, Faculty of Engineering, The University of Sydney, Sydney, Australia (e-mail: dacheng.tao@gmail.com).}
}
\maketitle


\begin{abstract}
The goal of aspect-based sentiment analysis (ABSA) is to determine the sentiment polarity towards an aspect. Because of the expensive and limited amounts of labelled data, the pretraining strategy has become the de facto standard for ABSA. However, there always exists a severe domain shift between the pretraining and downstream ABSA datasets, which hinders effective knowledge transfer when directly fine-tuning, making the downstream task suboptimal. To mitigate this domain shift, we introduce a unified alignment pretraining framework into the vanilla pretrain-finetune pipeline, that has both instance- and knowledge-level alignments. Specifically, we first devise a novel coarse-to-fine retrieval sampling approach to select target domain-related instances from the large-scale pretraining dataset, thus aligning the instances between pretraining and the target domains (\textit{First Stage}). Then, we introduce a knowledge guidance-based strategy to further bridge the domain gap at the knowledge level. In practice, we formulate the model pretrained on the sampled instances into a knowledge guidance model and a learner model. On the target dataset, we design an on-the-fly teacher-student joint fine-tuning approach to progressively transfer the knowledge from the knowledge guidance model to the learner model (\textit{Second Stage}). Therefore, the learner model can maintain more domain-invariant knowledge when learning new knowledge from the target dataset. In the \textit{Third Stage,} the learner model is finetuned to better adapt its learned knowledge to the target dataset. Extensive experiments and analyses on several ABSA benchmarks demonstrate the effectiveness and universality of our proposed pretraining framework. Our source code and models are publicly available at https://github.com/WHU-ZQH/UIKA.
\end{abstract}

\begin{IEEEkeywords}
Aspect-based sentiment analysis, pretraining, domain shift, instance alignment, knowledge alignment.
\end{IEEEkeywords}

%
\IEEEpeerreviewmaketitle

\section{Introduction}
\label{section1}
%
%
%
%
\IEEEPARstart{A}{S}, a fine-grained sentiment analysis task, aspect-based sentiment analysis (ABSA) has attracted much attention in the community of natural language processing (NLP). Specifically, ABSA contains several subtasks, such as aspect term sentiment classification and aspect extraction\emph{etc} \cite{liu2012survey,pontiki-etal-2014-semeval,tang2016aspect}. 
In this paper, we mainly denote ABSA with the aspect term, sentiment classification, which means it is formulated for determining the sentiment polarities of a sentence towards the given aspect terms.
 Note that the aspect terms are the target entities that appear in the sentences \cite{xue2018aspect,wang2022contrastive,chen2022retrieve}. For example, given a sentence ``The \emph{\textbf{food}} was good, but not at all worth the \emph{\textbf{price}}'', and the aspect terms of ``\emph{\textbf{food}}'' and ``\emph{\textbf{price}}'', the task of ABSA is  to determine the ``positive" and ``negative" sentiment polarities of the terms, respectively.

\begin{figure}[t]
	\centering
	\includegraphics[width=0.48\textwidth]{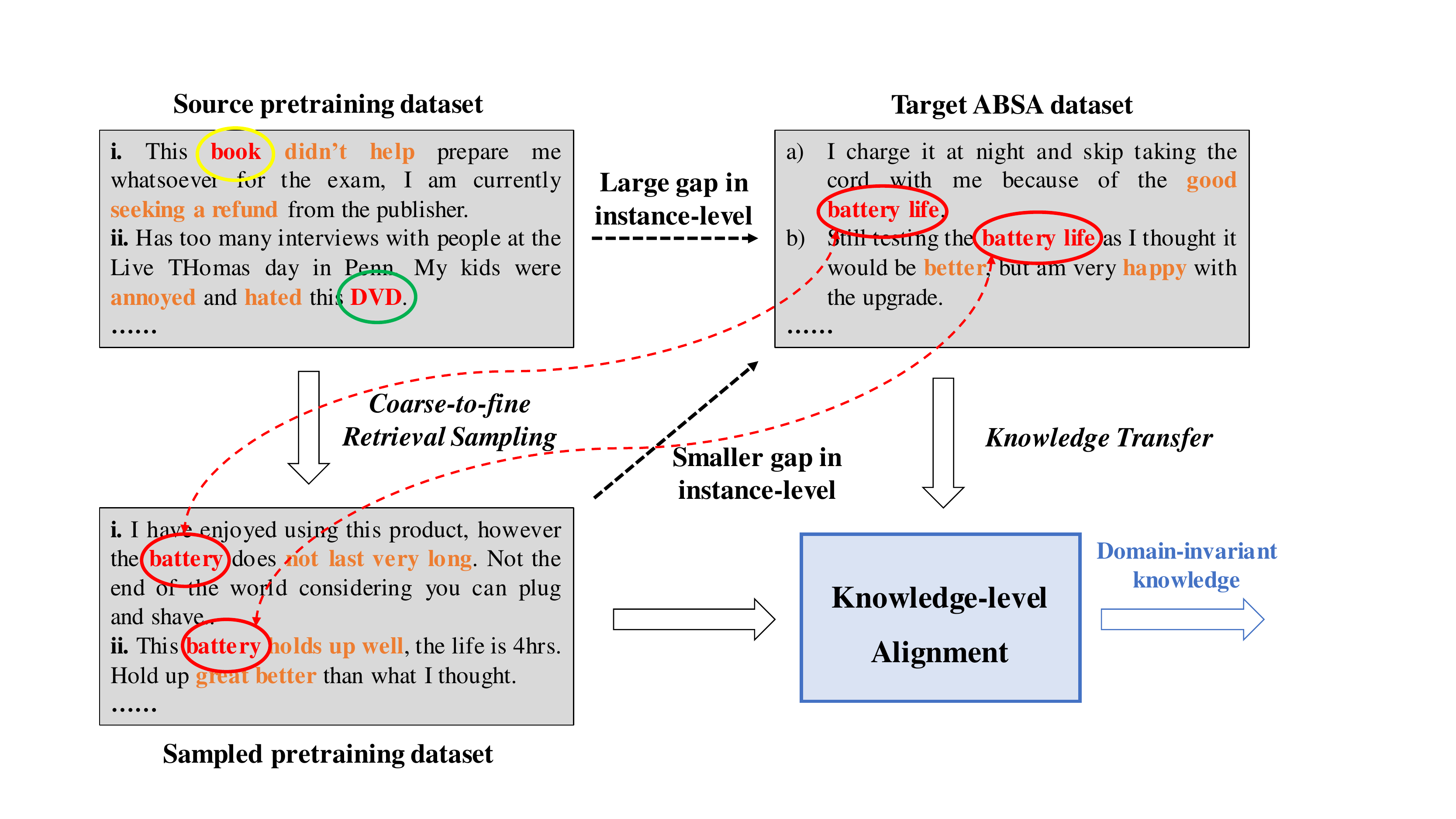} 
	\caption{The illustration of instance- and knowledge-level alignments, where instance alignment aims to align the instances of pretraining and the target datasets, and knowledge alignment aims to learn the domain-invariant knowledge. Notably, the words in the red and orange colours are the aspect terms and their related opinion words, respectively. The boxes in the same colour refer to the related aspect terms.}
	\label{fig0}
\end{figure}

The existing ABSA approaches are mainly based on deep neural networks (DNNs), including the LSTM-based methods \cite{wang2016attention,tang2020dependency}, CNN-based methods \cite{xue2018aspect,chen2020inducing} and syntax-based methods \cite{zhang-etal-2019-aspect, wang2020relational}. These methods commonly require full supervision from a large high-quality training corpus to achieve respectable performances. However, due to the difficulty of annotating aspect-level labels~\cite{zhang2022towards}, the current ABSA datasets are usually very small, making it difficult for the prevailing methods to fully exploit their model capacities and attain optimal performances. To ensure effective training on ABSA datasets with limited data, a pretraining strategy is usually adopted for DNNs.

For ABSA, the current pretraining strategy first trains DNNs on some larger-scale datasets of different tasks, such as sentence-level sentiment analysis datasets~\cite{che@taslp,tang2014learning,tang2015document}. Then, the pretrained DNNs are finetuned on the target ABSA dataset; thus, the knowledge learned from the large datasets can be transferred to the ABSA task. In practice, this pretraining strategy can help these ABSA methods achieve higher training efficiencies and better performances \cite{he2018exploiting,chen2019transfer}. Although promising progress has been witnessed, we find that most current pretraining-based methods tend to underestimate the severe domain shift problem between the pretraining and the downstream target datasets, where domain shift refers to the difference between the upstream and downstream datasets, \textit{e.g.}, product topic, text style, grammar and text granularity.
Taking a sentence-level pretraining dataset as an example, the pretraining dataset tends to be several orders of magnitude larger than the downstream ABSA dataset, and importantly, most are unrelated to the ABSA datasets~\cite{zhou2020position}. Such a domain gap will make the pretrained model deteriorate dramatically during fine-tuning, which has been proven in previous studies~\cite{howard-ruder-2018-universal, chen-etal-2020-recall, rusu2016progressive}.

To address the above problems, we propose a novel unified instance and knowledge alignment pretraining framework, namely, UIKA, which can gradually alleviate the domain shift problem. For better understanding, we show an illustration of the alignment procedure in Fig.~\ref{fig0}. Specifically, considering that the pretraining datasets usually contain large-scale instances, and most of them are unrelated to the target ABSA data, we first introduce a coarse-to-fine sampling approach to select the target domain-related instances. Afterwards, inspired by the studies \cite{gururangan2020don,xu2019self} that introduce continued pretraining and teacher-student learning strategies for tackling the issue of domain shift at the knowledge level, we further propose to introduce a knowledge guidance-based strategy to progressively align the knowledge between pretraining and the target domains. In practice, we train the model on the sampled pretraining instances and attempt to describe the pretrained model with a knowledge guidance model and a learner model, both of which initially share the same model definitions and the same prior knowledge. Then, we aggressively finetune the knowledge guidance model on the target ABSA dataset. Meanwhile, the knowledge guidance model will guide the training of the learner model, which is updated with a more conservative pace. Accordingly, although the knowledge guidance model likely will forget the prior knowledge due to the significant domain gap, it is highly possible that the domain-invariant knowledge can be better maintained in the learner model. Subsequently, by further finetuning the learner model on the ABSA dataset, it can achieve a more promising performance.

This paper's main research contributions can be summarized as follows:
\begin{enumerate}
	\item We propose a novel unified instance and knowledge alignment pretraining (UIKA) framework to recast the vanilla pretraining strategy in ABSA. The UIKA introduces a coarse-to-fine retrieval sampling strategy and a knowledge guidance-based training scheme to effectively alleviate the domain shift at the instance and knowledge levels, respectively.
	
	\item Our proposed UIKA framework is easy to implement, is model independent and does not require any complicated model modifications, making it possible to apply it to a broader range of DNN-based ABSA models. 
	
	\item 
	Extensive experimental results demonstrate that the UIKA framework significantly and consistently outperforms the vanilla pretraining strategy and achieves SOTA level performances on most of the evaluated benchmarks.
\end{enumerate}

The rest of this paper is organized as follows. In Sec.~\ref{section2}, we briefly review the related works. In Sec.~\ref{section3}, we introduce our proposed framework in detail. Sec.~\ref{section4} reports and discusses our experimental results, and is followed by a discussion in Sec.~\ref{section5}. Finally, we conclude our study in Sec.~\ref{section6}.

%

\begin{figure*}[t]
	\centering
	\includegraphics[width=0.90\textwidth]{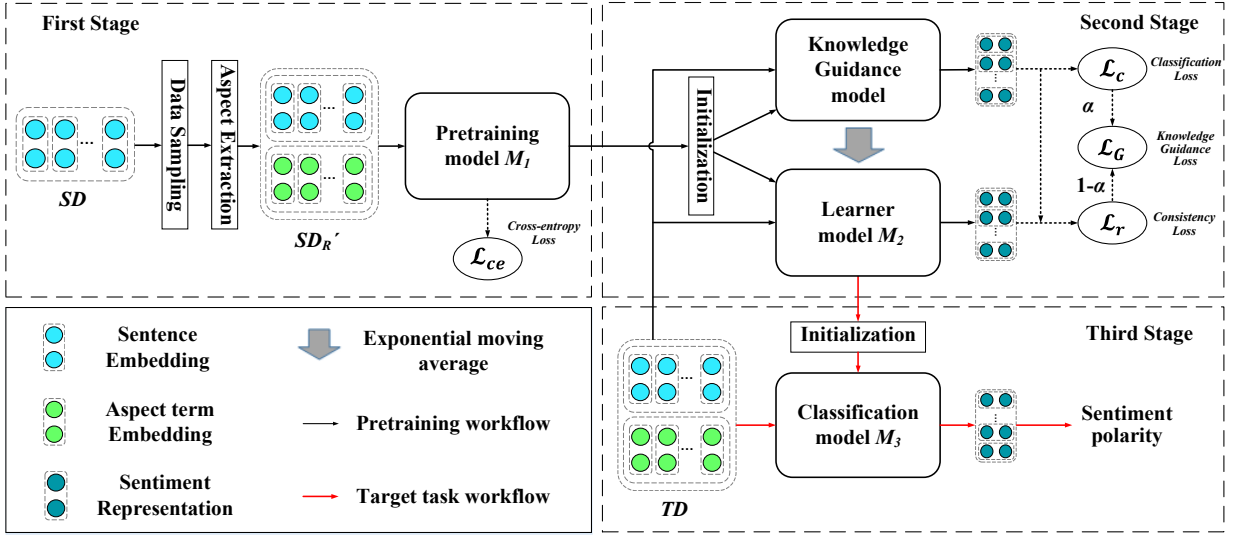} 
	\caption{The architecture of our UIKA pretraining framework contains three training stages to alleviate the domain shift between the pretraining and target ABSA domains. Specifically, the first stage performs instance-level domain adaptation, while the second stage further bridges the domain gap at the knowledge level. Notably, $SD$, $SD_R'$ and $TD$ denote the sentence-level, pseudosampled aspect-level and target aspect-level datasets, respectively.}
	\label{fig1}
\end{figure*}

\section{Related Works}
\label{section2}
\subsection{Aspect-based Sentiment Analysis}
In recent years, DNNs have attracted increasing attention in the ABSA research community because they are much more powerful than hand-crafted features and achieve superior performances. The DNN-based methods for ABSA can be roughly classified into three categories, \textit{i.e.} LSTM-based methods \cite{tang2016effective,wang2016attention,ma2017interactive}, CNN-based methods \cite{xue2018aspect,li2018transformation,chen2020inducing} and syntax-based methods \cite{tang2020dependency,wang2020relational,li-etal-2021-dual-graph,pang-etal-2021-dynamic,zhong2023KGAN}. 

First, owing to the ability to learn sequential patterns, using LSTMs to extract aspect-specific features has become the mainstream approach for ABSA. Tang \textit{et al.} \cite{tang2016effective} proposed a target-dependent LSTM (TD-LSTM) to capture aspect-related information. Wang \textit{et al.} \cite{wang2016attention} improved upon the TD-LSTM by introducing an attention mechanism to explore the potential correlations between aspects and opinion words. Additionally, in the study of \cite{ma2017interactive}, two LSTMs were used to encode the context and aspect terms, and then an interactive attention mechanism was introduced to extract the more relevant information between the context and aspect features. 

Later, considering the complexity and inefficiency of LSTMs, many models attempted to employ more efficient CNNs to capture the compositional structure and n-gram features. Xue and Li \cite{xue2018aspect} proposed a gated convolution network to extract the contextual features and employed the gate mechanism to selectively output the final sentiment features. In CNN-based methods, it is common to employ the average of aspect embeddings as the aspect representation, which causes a loss of sequence information. To address this problem, Li \textit{et al.} \cite{li2018transformation} introduced a target-specific transformation component based on CNNs to better learn the target-specific representation.

Recently, to tackle the multiple-aspect problem, many studies have focused on explicitly leveraging the syntactic structure to establish an aspect and opinion word connection. Specifically, these methods introduced dependency parse trees to incorporate the syntactic information. Zhang \textit{et al.} \cite{zhang-etal-2019-aspect} first utilized dependency trees to represent sentences, and then proposed graph convolution networks (GCNs) to exploit the syntactical information from dependency trees. Additionally, to better syntactically connect aspect and opinion words, Wang \textit{et al.} \cite{wang2020relational} presented a novel aspect-oriented dependency tree structure and employed a relational graph attention network to encode the tree structure. 

These DNN-based methods have achieved some promising results, however, some challenges still remain in the ABSA task, \emph{e.g.,} the lack of labelled data. It is usually difficult to obtain a 
well-performing model without a large set of labelled data.
To this end, recent studies focus on effectively migrating the knowledge encoded on the pretrained models into the downstream tasks, \textit{e.g.} ABSA.

\subsection{Pretraining for ABSA}
In the NLP community, many methods that use pretraining strategies have made impressive achievements, such as the representative word embedding models (\textit{e.g.} Word2vec \cite{mikolov2013distributed} and GloVe \cite{pennington2014glove}) and the contextualized pretrained language model, denoted as PLM, (\textit{e.g.} GPT \cite{radford2018improving} and BERT \cite{devlin2018bert}). These general pretrained models focus on learning descent word embeddings or contextual representations, which contain rich semantic features. Using these powerful pretrained models could improve the performance of downstream tasks by a large margin, especially in low- and medium-resource scenarios. 
\zqh{In the ABSA field, there are also many PLM-based models~\cite{wang2020relational,li2020enhancing,mao2021joint,zhang2022ssegcn,gao2022lego,cao2022aspect,chen2022discrete,yan2021unified}, such as R-GAT-BERT~\cite{wang2020relational} and KGAN-BERT~\cite{zhong2023KGAN} }

Additionally, since it is usually time-consuming to train and finetune these general pretrained models, some works have focused on studying task-specific pretraining. Specifically, task-specific pretraining attempts to pretrain the model from a task-related corpus, and then finetune the pretrained model on the target-task corpus to transfer the knowledge. In the ABSA task, we witnessed many studies on task-specific pretraining \cite{he2018exploiting,chen2019transfer,li2019exploiting,zhou2020position,chen-etal-2020-recall,xu2019bert,liang2021iterative}. He \textit{et al.} \cite{he2018exploiting} proposed a transfer learning method that combined pretraining and multitask learning to exploit the knowledge gained from the sentence-level corpus for ABSA. In the study of Chen~\textit{et al.}~\cite{chen2019transfer}, the proposed TransCap also attempted to transfer sentence-level knowledge to the aspect-level dataset. To reduce the noise of pretraining the datasets, Zhou~\textit{et al.}~\cite{zhou2020position} introduced three sampling approaches to select the pretraining instances and transferred the knowledge into ABSA from four different levels.

The above task-specific pretraining-based methods have achieved remarkable performances in the ABSA field , however, only a few methods \cite{rusu2016progressive,li2019exploiting} have recognized the negative influence of the domain shift between the pretraining and the target datasets. Li~\textit{et al.}\cite{li2019exploiting} proposed a coarse-to-fine task transfer network to reduce the domain shift between different tasks and datasets. In the work~\cite{rusu2016progressive}, a progressive network approach is proposed to leverage prior knowledge, and to alleviate forgetting via lateral connections to previously learned features. Notably, our motivation is similar to that in \cite{rusu2016progressive} but our proposed method is significantly different from the progressive networks approach. To be more specific, instead of directly concatenating the features of the pretrained model and target model, we use a knowledge guidance-based strategy to encourage the pretrained knowledge model to guide the learning of the target learner model. \zqh{The knowledge guidance-based strategy can also be viewed as a knowledge distillation (KD)~\cite{hinton2015distilling} technology, which attempts to leverage the teacher model to guide the training of students. The KD has the powerful ability to transfer knowledge, which has been proven effective in many fields~\cite{ding2021understanding,hahn2019self}.
}

Additionally, there are also two differences between our method and prior works \cite{li2019exploiting,rusu2016progressive,zhou2020position}. On the one hand, we bridge the domain gap, step-by-step, via both instance-level and knowledge-level alignments, while Li~\textit{et al.}~\cite{li2019exploiting} and Rusu~\textit{et al.}~\cite{rusu2016progressive} only focus on the knowledge level, and \zqh{and Zhou~\textit{et al.}~\cite{zhou2020position} mainly use coarse-grained instance-level alignments}. On the other hand, the method in \cite{li2019exploiting} is only suitable for attention-based RNN models, whereas in this paper, we propose a pretraining framework that can be adapted to any DNN.

\section{Unified Instance and Knowledge Alignment Pretraining}
\label{section3}
\subsection{Problem Formulation}	
Suppose we have a source sentence-level dataset, denoted as $SD$, and a target aspect-level dataset, denoted as $TD$. We then denote a sentence as $S\in TD$ and the related aspect term as $T$. Notably, $T$ is usually a subsequence of $S$, and $S$ could contain multiple $T$ with different polarities. The goal of ABSA is to predict the sentiment polarity $y\in \{0, 1, 2\}$ of the sentence $S$ towards a specific aspect term $T$, where 0, 1 and 2 represent the \emph{positive}, \emph{neutral} and \emph{negative} polarities, respectively. The UIKA aims to improve the performance of the predictive model for the ABSA task using the semantic knowledge learned from $SD$.

\subsection{Multistage Pretraining Procedure}
The overall architecture of our proposed UIKA framework is shown in Fig.~\ref{fig1}. In general, the UIKA framework consists of three processing stages. The first stage of our pretraining framework introduces a novel sampling strategy to select the target domain-related instances from $SD$ and employs an ABSA DNN to learn richer knowledge from the sampled instances. Then, in the second stage, we formulate the DNN model pretrained in the first stage into a knowledge guidance model and a learner model. Both models are initialized with the weights learned in the first stage before adapting their knowledge to $TD$.
During the training procedure of the second stage, we make the knowledge guidance model learn more quickly and guide the training of the learner model to learn the domain-invariant knowledge. Finally, in the third stage, we finetune the learner DNN model trained in the second stage on $TD$ to obtain a more promising performance.

\paragraph{Initial Sampling-based Pretraining Stage} 
\label{section3.2.1}

For the first stage, we perform vanilla pretraining for an ABSA DNN model on $SD$ to gain rich prior knowledge about the semantics. As stated above, $SD$ is usually large-scale and contains a considerable number of various instances. Pretraining on all instances of $SD$ is time-consuming and introduces considerable noise. Therefore, we devised a novel coarse-to-fine retrieval strategy to sample the target domain-related instances from $SD$. Notably, for a given target instance, an intuitive approach is to map all the source instances and the target instance to a shared semantic space, and then retrieve the optimal instances via a metric in one step. However, we find that such a process is costly and redundant, as most source instances are useless. Hence, we attempt to retrieve the required instances in a coarse-to-fine manner, \textit{i.e.}, first filtering the obviously dissimilar samples and then selecting the more related ones from the filtered samples.

In practice, we first treat the sentence $S$ of $TD$ as a query and employ an efficient relevance-based retrieval approach\footnote{Note that we employ such a simple and classical approach, owing to its generality and rapidity. We state that it will be more accurate when using a sophisticated approach, \textit{e.g. }, using the cosine similarity of sentence embedding as the metric to retrieve the related sentences via a pretrained language model, however this is more time-consuming and costly.} (\textit{i.e.} BM25 \cite{robertson2009probabilistic}) to select the top-\textit{N}-related instances (namely, $C_{result}$), where \textit{N} is the number of sampled instances for each query. These sampling instances are then encoded into a semantic vector space, where the larger cosine similarity between the two instances refers to the higher semantic similarity. Next, to further select the related instance, we follow Gururangan~\textit{et al.}~\cite{gururangan2020don}, calculate the cosine similarity between $S$ and each instance of $C_{result}$, and select the top-\textit{K} semantically similar instances as fine-grained sampling instances (namely, $F_{result}$). Note that \textit{K} is the number of fine-grained sampled instances and is less than \textit{N}. As a result, we finally collect all the related instances and obtain a new pretraining dataset (namely, $SD_R$)\footnote{\zqh{The data volume of $SD_R$ depends on the number of samples in the target ABSA datasets. Taking Laptop14 (approximately 3,000 samples) as an example, we collect approximately 3,000*\textit{K} data points.}}. For a better understanding, we present the operation of the sampling strategy in Algorithm~\ref{Algorithm 1}.

\begin{algorithm}[t]
	\caption{Coarse-to-fine sampling strategy.} 
	\label{Algorithm 1}
	\LinesNumbered
	\KwIn{\\\qquad \textbf{$SD$}: the source pretraining dataset;\\
		\qquad \textbf{$TD$}: the target ABSA dataset;\\
		\qquad \textbf{$N$}: the number of coarse-grained sampling instances for each query; \\
		\qquad \textbf{$K$}: the number of fine-grained sampling instances for each query}
	\KwOut{\\\qquad $SD_R$: the sampled pretraining dataset }
	\For{each sentence $S$ $\in$ $TD$}{	
		\tcc{\textit{Coarse-grained Sampling Stage}}
		Let $C_{score}$, $C_{result}$ be new arrays \\
		$C_{score} \gets$BM25$(S, SD)$  \\
		$C_{result} \gets$GetTop$(C_{score}, SD, N)$  \\
		\tcc{\textit{Fine-grained Sampling Stage}}
		Let $Score \gets$ 0, $F_{score}$, $F_{result}$ be new arrays, and $S_{emb}$, $I_{emb}$ be new matrices \\
		$S_{emb} \gets$GetEmbedding$(S)$ \\
		\For{each instance $I$ $\in$ $C_{result}$}{
			$I_{emb} \gets$GetEmbedding$(I)$  \\
			$Score \gets$Similarity$(S_{emb}, I_{emb})$   \\
			$F_{score}.append(Score)$   \\
		}
		$F_{result} \gets$ GetTop$(F_{score}, C_{result}, K)$ \\
		$SD_R.append(F_{result})$   \\
	}
	return $SD_R$;
\end{algorithm}

Additionally, to make the pretrained DNN model more effectively learn the aspect term-related knowledge, we convert the sampled dataset $SD_R$ to a pseudo aspect-level dataset $SD_R'$ that is more sensitive to the aspect terms. Notably, the source sentence-level pretraining datasets used for sentiment analysis are generally review datasets (\textit{e.g.} Amazon and Yelp datasets), in which the aspect terms are usually nouns or noun phrases \cite{Hu2004}. Therefore, we follow Hu \emph{et al.} \cite{Hu2004} and extract nouns or noun phrases as aspect terms via the part-of-speech tagging and extra preprocessing of the words.\footnote{We employ the approach due to its simplicity. Note that the more advanced approaches could improve the performance but this is not our focus in the paper.} More specifically, the detailed operation of aspect extraction can be found in Sec.~\ref{section 5.1}.

By pretraining a DNN model with $SD_R'$, we can obtain the initial pretrained model, denoted as $M_1$, for the first stage. Let $y_{s_i}$ be the label of the $i$-th instance from $SD_R'$ and $p_{s_i}$ be the predicted polarity given by the classification layer of $M_1$ for the $i$-th data example. We apply the cross-entropy loss function for $M_1$ to train with $SD_R'$ as follows:
\begin{equation}
\mathcal{L}_{ce}=-\sum_i \sum_j  y^j_{s_i} \log( p^j_{s_i})
\end{equation}
where $i$ indexes the data examples in $SD_R'$ and $j$ indexes the sentiment classes.

\paragraph{Knowledge Guidance-based Pretraining Stage} After the first stage, we obtain $M_1$, which encodes the abundant semantic knowledge learned from $SD_R'$. Although the sampling strategy in the first stage can select the target domain-related instances and alleviate the domain shift at the instance level, there is still some domain shift between $SD_R'$ and $TD$, which would hinder the knowledge transfer performance. Moreover, by directly finetuning the weights of $M_1$ on the ABSA dataset $TD$ runs the risk of making pretrained $M_1$ quickly ``forget'' its learned knowledge from $SD_R'$ \cite{chen-etal-2020-recall}. Therefore, rather than directly feeding prior knowledge into the DNN models for training on the $TD$, we introduce an on-the-fly teacher-student joint finetuning approach to further bridge the domain gap at the knowledge level. This learning procedure forms the second pretraining stage in our UIKA. 

The on-the-fly teacher-student approach represents the $M_1$ with two models that share a consistent architecture, \textit{i.e.} a knowledge guidance model and a learner model. After initializing both models with weights from $M_1$, the knowledge guidance model aims to intensively learn new knowledge from $TD$. Meanwhile, we make the learner model follow the learning of the knowledge guidance model for updating its own representation. 

As shown in the second stage of Fig.~\ref{fig1}, we first initialize the knowledge guidance model and the learner model with the same architecture and same parameters learned within $M_1$. Then, we feed the target aspect-level data $TD$ to both the knowledge guidance model and learner model to conduct the second pretraining stage. The supervision of $TD$ is only used in the knowledge guidance model. In addition, we introduce a knowledge guidance loss $\mathcal{L}_{G}$ to implement the proposed knowledge guidance-based pretraining.
We define the knowledge guidance loss $\mathcal{L}_{G}$ as follows:
\begin{equation} \label{eq2}
\mathcal{L}_{G} =\alpha(e)  * \mathcal{L}_{c} + (1-\alpha(e) )*\mathcal{L}_{r}
\end{equation}
where $\mathcal{L}_{c}$ is the classification loss and $\mathcal{L}_{r}$ is the representation consistency loss. Suppose $p_{g_i}$ is the prediction for the $i$-th dataset made by the knowledge guidance model, and $p_{l_i}$ is the prediction for the same dataset made by the learner model. $\mathcal{L}_{c}$ and $\mathcal{L}_{r}$ can be, respectively defined as follows:
\begin{align}
\mathcal{L}_{c} &=-\sum_i \sum_j  y^j_{t_i} \log( p^j_{g_i})   \\
\mathcal{L}_{r} &=\sum_i \sum_j (p^j_{g_i}- p^j_{l_i})^2
\end{align}
where $y^j_{t_i}$ is the label for the $i$-th data example from the target $TD$, and $j$ indexes the different sentiment classes.

Additionally, the $\alpha(e)$ of Eq.\ref{eq2} represents an annealing function that controls the degree of knowledge transfer. Note that the annealing function could be various functions, e.g., the sigmoid function or linear function. In this study, we empirically employ the linear function as the annealing function $\alpha(e)$ since the sigmoid function \cite{chen-etal-2020-recall} causes the problem of unstable updates and hinders the knowledge transfer performance. More specifically, $\alpha(e)$ is calculated as follows:
	\begin{equation} \label{eq5}
	\alpha(e) =1-\dfrac{e}{E+1}
	\end{equation}
where $e$ denotes the index of the training epoch, and $E$ denotes the number of all training epochs, which is empirically set as 10. Note that the analysis of $\alpha(e)$ can be found in Sec.~\ref{section4.4}.

Afterwards, when the parameters of the knowledge guidance model (namely, $\theta_{g}$) are updated according to $\mathcal{L}_{G}$, we update the parameters of the learner model (namely, $\theta_l$) using an exponential moving average method \cite{10.5555/3294771.3294885}, which can be defined as follows:
\begin{equation} \label{eq3}
\theta_l^t=\beta * \theta_l^{t-1} + (1-\beta)*\theta_{g}^t
\end{equation}
where $t$ indexes the training iterations and $\beta$ is a smoothing coefficient hyperparameter. Note that we do not update the learner model \textit{w.r.t.} $\mathcal{L}_G$.

\begin{figure}[tp]
	\centering
	\includegraphics[width=0.48\textwidth]{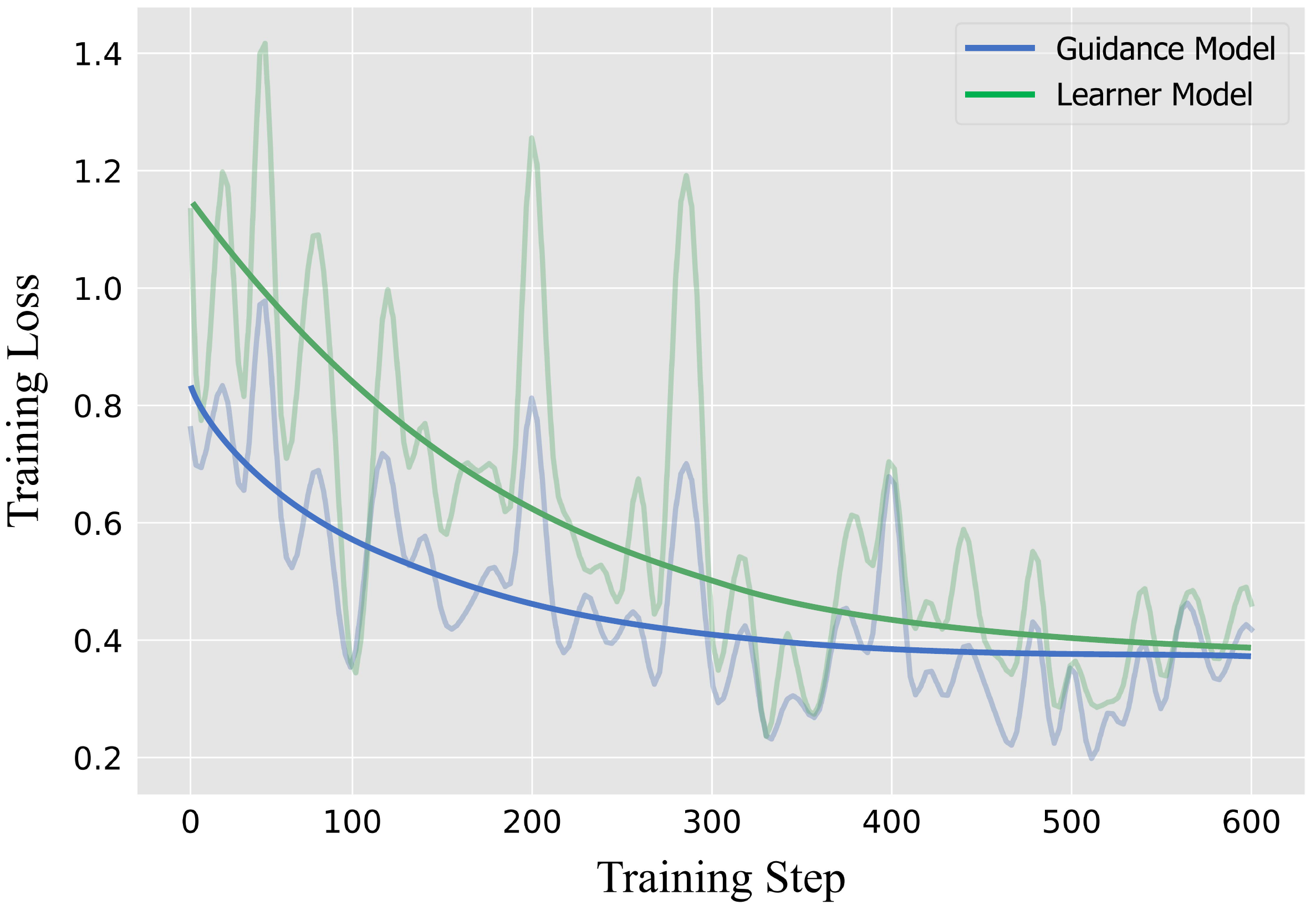} 
	\caption{\zqh{Training curve of the RGAT model in the knowledge guidance stage. Note that we use Restaurant14 as the target dataset here. }}
	\label{fig7}
\end{figure}

According to the procedure described between Eq.~\ref{eq2} and Eq.~\ref{eq3}, the learner model follows the update of the knowledge guidance model but at a slower pace, which is controlled by the value of $\beta$. Different from the vanilla finetuning, which directly optimizes the learner model with a classification loss of the ABSA task, its weights are only updated to be similar to the knowledge guidance model according to Eq.~\ref{eq3}. In this way, even if the knowledge guidance model may easily forget pretrained knowledge through fine-tuning on the ABSA dataset, the learner model can still preserve the important knowledge learned from $M_1$. \zqh{To have a closer look, we illustrate the training curves of both models (in this stage) in Fig.~\ref{fig7}. The update rate of the learning model is much slower than that of the guidance model. That is, the learning model has a smoother knowledge-learning process and has less risk of forgetting knowledge.} Meanwhile, the representation consistency loss $\mathcal{L}_{r}$ can help ensure that the knowledge guidance model will not deviate too much from the learner model during this pretraining stage so that it can provide meaningful information to guide the training of the learner model.

\paragraph{Final Finetuning Stage} After the second stage, we obtain the trained learner model and formulate it as the pretrained model $M_2$. In this stage, we employ the ABSA model as the final classifier, which has the same architecture as the learner model except for the position module (the implementation of UIKA on different baseline models is described in Section~\ref{section3.3}). To train this classifier on the target dataset, we initialize its parameters using $M_2$ and then feed $TD$ to it, obtaining the required classification model $M_3$.

According to the study \cite{howard-ruder-2018-universal}, the prior learned knowledge would be forgotten if the domain gap between the source and target datasets is large, however, it could be maintained if both domains share a certain similarity. In our framework, the first pretraining stage samples the target-related instances and aligns the source and target domains at the instance level, while the second pretraining stage aligns the knowledge and learns the domain-invariant knowledge through the knowledge guidance procedure. As a result, the proposed unified alignment pretraining framework can, step-by-step, help alleviate the domain gap between the source pretraining and target ABSA datasets; therefore, the prior knowledge used in the third stage would not be severely forgotten during fine-tuning. Overall, our proposed pretraining strategy can achieve a better transfer of the pretrained knowledge to the target ABSA model.


\begin{table}[t]
	\centering
	\caption{Statistics of the evaluated aspect-level datasets.}
	\label{table0}
	\begin{tabular}{lcccc}
		\Xhline{1.2pt}
		\textbf{Datasets}&\textbf{Division}  & \textbf{\#Positive} & \textbf{\#Negative} & \textbf{\#Neutral} \\
		\cmidrule(){1-5}\morecmidrules\cmidrule(){1-5}
		\multirow{2}{*}{Laptop14~\cite{pontiki-etal-2014-semeval}} & Train   & 980       & 858       & 454      \\ 
		& Test    & 340       & 128       & 171      \\
		\midrule
		\multirow{2}{*}{Restaurant14~\cite{pontiki-etal-2014-semeval}}   & Train   & 2,159      & 800       & 632      \\
		& Test    & 730       & 195       & 196      \\
		\midrule 
		\multirow{2}{*}{Twitter~\cite{dong2014adaptive}}  & Train   & 1,567      & 1,563      & 3,127     \\
		& Test    & 174       & 174       & 346    \\ \midrule
		\multirow{2}{*}{\zqh{Restaurant15~\cite{pontiki2015semeval}}}   & \zqh{Train}   & \zqh{912}      & \zqh{256}       & \zqh{36}      \\ 
		& \zqh{Test}    & \zqh{326}       & \zqh{182}       & \zqh{34}      \\ \midrule
		\multirow{2}{*}{\zqh{Restaurant16~\cite{pontiki2016semeval}}}   & \zqh{Train}   & \zqh{1,240}     & \zqh{439}       & \zqh{69}      \\
		& \zqh{Test}    & \zqh{469}       & \zqh{117}       & \zqh{30}      \\ \midrule
		\multirow{3}{*}{\zqh{MAMS~\cite{jiang2019challenge}}}   & \zqh{Train}   & \zqh{3,380}     & \zqh{2,764}       & \zqh{5,042}      \\
		& \zqh{Dev}    & \zqh{403}       & \zqh{325}       & \zqh{604}      \\
		& \zqh{Test}    & \zqh{400}       & \zqh{329}       & \zqh{607}      \\
		\Xhline{1.2pt}
	\end{tabular}
\end{table}

\begin{table*}[t]
	\centering
	\caption{\zqh{Comparison with previous work on several SentEval-based datasets. Most results are retrieved from corresponding papers. ``$\dag$'' and ``$\ddag$'' indicate the results are collected from \cite{zhou2020sk}, \cite{wang2020relational}, respectively. ``-'' means the results are not reported in their original works and ``$\star$'' denotes the results that we reproduced. The best results for each pretrained model are in bold, while the second-best results are underlined.} Following the previous work~\cite{zhou2020sk}, we apply a student t-test~\cite{mendenhall2012introduction} to calculate the statistical significance, and the results show that  UIKA significantly outperforms baselines on these benchmarks at the significance level of $p < 0.05$.}
	\label{table1}
	\begin{tabular}{llcccccccccc}
		\Xhline{1.2pt}
		\multirow{2}{*}{\textbf{Embedding}} & \multirow{2}{*}{\textbf{Method}}           & \multicolumn{2}{c}{\textbf{Laptop14}}                  & \multicolumn{2}{c}{\textbf{Restaurant14}}    & \multicolumn{2}{c}{\textbf{Twitter}}                   & \multicolumn{2}{c}{\zqh{\textbf{Restaurant15}}} & \multicolumn{2}{c}{\zqh{\textbf{Restaurant16}}} \\   \cmidrule(lr){3-12}
		&  & Acc.  & F1.  & Acc.  & F1. & Acc. & F1.  & \zqh{Acc.}  & \zqh{F1.}   & \zqh{Acc.}    & \zqh{F1.}   \\ \cmidrule(lr){1-12}\morecmidrules\cmidrule(lr){1-12}
		\multirow{25}{*}{GloVe}                                                     & TD-LSTM$\dag$ \cite{tang2016effective}  & 71.83     & 68.43       & 78.00       & 68.43     & 70.80       & 69.00  &\zqh{-}  &\zqh{-} &\zqh{-} &\zqh{-}        \\ 
		& IAN$\dag$ \cite{ma2017interactive} & 71.79       & 65.92      & 77.86       & 66.31           &71.82          &69.11   &\zqh{-} &\zqh{-} &\zqh{-} &\zqh{-}         \\      
        & \zqh{ATAE-LSTM$^\dag$ \cite{wang2016attention}}     &\zqh{70.66}   &   \zqh{64.95}      & \zqh{76.44}  &\zqh{64.46}     &\zqh{74.83}    &\zqh{74.04}     & \zqh{78.48}   & \zqh{62.84}   & \zqh{83.77}   & \zqh{61.71}     \\
		& \zqh{GCAE \cite{xue2018aspect}}   & \zqh{73.56}      & 67.84       & \zqh{79.27}      & \zqh{67.66}     & \zqh{77.69}       & \zqh{76.58}  &\zqh{-} &\zqh{-} &\zqh{-} &\zqh{-}     \\
		& \zqh{RAM \cite{chen2017recurrent}}     & \zqh{74.49}    & \zqh{71.35}   & \zqh{80.23}     & \zqh{70.80}  & \zqh{69.36}    & \zqh{67.30}      & \zqh{79.98}   & \zqh{60.57}          & \zqh{83.88}  & \zqh{62.14}   \\
		& TransCap \cite{chen2019transfer} & 73.87      & 70.10     & 79.29      & 70.85          & -          & -     &\zqh{-} &\zqh{-} &\zqh{-} &\zqh{-}      \\
		& TNet-AS \cite{li2018transformation}      & 76.54      & 71.75   & 80.69    & 71.27    & 74.97     & 73.60    & \zqh{78.47}    & \zqh{59.47}   & \zqh{89.07}   & \zqh{70.43}    \\
		& MGAN \cite{li2019exploiting}    & 76.21  & 71.42   & 81.49     & 71.48    & 74.62     & 73.53   & \zqh{-}     & \zqh{-}     & \zqh{-}   & \zqh{-}   \\
		& TNet-ATT \cite{tang2019progressive}  & 77.62    & 73.84    & 81.53          & 72.90     & 77.72    & 78.61  &\zqh{-} &\zqh{-} &\zqh{-} &\zqh{-}   \\
		& MCRF-SA \cite{xu-etal-2020-aspect}   & 77.64  & 74.23     & 82.86     & 73.78  & -    & -     & \zqh{80.82} & \zqh{61.59}    & \zqh{89.51}   & \zqh{\textbf{75.92}}   \\ 
		& \zqh{ASCNN \cite{zhang-etal-2019-aspect}}    & \zqh{72.62}   & \zqh{66.72}     & \zqh{81.73}    & \zqh{73.10}   & \zqh{71.05}   & \zqh{69.45}      & \zqh{78.48}   & \zqh{58.90}   & \zqh{87.39} & \zqh{64.56}  \\
		& \zqh{ASGCN \cite{zhang-etal-2019-aspect}}    & \zqh{75.55}    & \zqh{71.05}     & \zqh{80.86}       & \zqh{72.19}       & \zqh{72.15}     & \zqh{70.40}    & \zqh{79.34}           & \zqh{60.78}          & \zqh{88.69}   & \zqh{66.64}   \\
		& \zqh{RACL \cite{chen2020relation}}    & \zqh{-}    & \zqh{71.09}     & \zqh{-}       & \zqh{74.46}    & \zqh{-}   & \zqh{-}    & \zqh{-}     & \zqh{68.69}    & \zqh{-}           & \zqh{-}            \\
		& DGEDT \cite{tang2020dependency}   & 76.80     & 72.30   & 83.90    & 75.10   & 74.80   & 73.40     & \zqh{82.10}           & \zqh{65.90}          & \zqh{90.80}   & \zqh{73.80}  \\
		& R-GAT~\cite{wang2020relational}     & 77.42   & 73.76    & 83.30    & 76.08  & 75.57        & 73.82      & \zqh{80.83}           & \zqh{64.17}          & \zqh{88.92}   & \zqh{70.89}   \\
		& DM-GCN \cite{pang-etal-2021-dynamic}   & 78.48    & 74.90    & 83.98     & 75.59    & 76.93   & 75.90  & \zqh{-}      & \zqh{-}   & \zqh{-}       & \zqh{-}       \\
		& Sentic-LSTM$^\dag$ \cite{ma2018sentic}   & 70.88   & 67.19     & 79.43     & 70.32  & 70.66       & 67.87     & \zqh{79.55}           & \zqh{60.56}          & \zqh{83.01}   & \zqh{68.22}    \\
		& Sentic GCN \cite{liang2022aspect}    & 77.90       & 74.71      & 84.03     & 75.38     & -           & -        & \zqh{82.84}           & \zqh{67.32}          & \zqh{\underline{90.88}}   & \zqh{\underline{75.91}}     \\
		& \zqh{SSEGCN \cite{zhang2022ssegcn}}    & \zqh{\textbf{79.43}}       & \zqh{\textbf{76.49}}      & \zqh{84.72}     & \zqh{77.51}     & \zqh{76.51}           & \zqh{75.32}       & \zqh{-}           & \zqh{-}          & \zqh{-}   & \zqh{-}     \\
         \cmidrule(lr){2-12}
		 & \zqh{RGAT~\cite{bai2021rgat}}     & \zqh{78.02}     & \zqh{74.00}   & \zqh{83.55}   & \zqh{75.99} & \zqh{75.36}    & \zqh{74.15}       & \zqh{80.33$^{\star}$}               & \zqh{61.10$^{\star}$}              & \zqh{88.41$^{\star}$}       & \zqh{62.79$^{\star}$}       \\
		&\multirow{1}{*}{\zqh{\textbf{RGAT+UIKA (Ours)}}} &\zqh{78.80} &\zqh{74.20} &\zqh{84.60} &\zqh{77.56} &\zqh{76.28} &\zqh{75.04} &\zqh{81.07} &\zqh{61.97} &\zqh{89.22} &\zqh{66.27} \\		
		\cdashline{2-12}
		& DualGCN \cite{li-etal-2021-dual-graph}    & 78.48    & 74.74   & 84.27 & \underline{78.08}    & 75.92    & 74.29    & 79.11$^{\star}$      & 62.25$^{\star}$   & 87.80$^{\star}$      & 70.34$^{\star}$    \\ 
		&\multirow{1}{*}{\textbf{DualGCN+UIKA (Ours)}}    &78.89    & 75.14    & \underline{85.19}   & \textbf{79.05} & 76.66    & 74.90     & 81.16     & 65.26   & 88.91      & 74.25   \\  		
		\cdashline{2-12}
		& \zqh{KGAN~\cite{zhong2023KGAN}}    & \zqh{78.91}     & \zqh{75.21}   & \zqh{84.46}    & \zqh{77.47} & \zqh{\underline{78.55}}      & \zqh{\underline{77.45}}       & \zqh{\underline{83.09}}           & \zqh{\underline{67.90}}          & \zqh{89.78}   & \zqh{74.58}      \\ 
		&\multirow{1}{*}{\zqh{\textbf{KGAN+UIKA (Ours)}}} &\zqh{\underline{79.31}} &\zqh{\underline{75.53}} &\zqh{\textbf{85.53}} &\zqh{78.00} &\zqh{\textbf{79.32}}  &\zqh{\textbf{77.95}}  &\zqh{\textbf{83.89}}  &\zqh{\textbf{68.52}} &\zqh{\textbf{90.92}} &\zqh{75.74}  \\		
		\midrule
		\multirow{17}{*}{BERT}      
        & \zqh{Vanilla BERT$^\ddag$ \cite{devlin2018bert}} & \zqh{77.58}     & \zqh{72.38}    & \zqh{85.62}       & \zqh{78.28}       & \zqh{75.28}   & \zqh{74.11}     & \zqh{83.40}    & \zqh{65.28}    & \zqh{89.54}   & \zqh{70.47}     \\
		& \zqh{R-GAT-BERT \cite{wang2020relational}}  & \zqh{78.21}    & \zqh{74.07}     & \zqh{86.60}     & \zqh{81.35}     & \zqh{76.15}   & \zqh{74.88}      & \zqh{83.22}           & \zqh{69.73}          & \zqh{89.71}   & \zqh{76.62}   \\
		& \zqh{RACL-BERT~\cite{chen2020relation}} & \zqh{-}     & \zqh{73.91}    & \zqh{-}       & \zqh{81.61}       & \zqh{-}   & \zqh{-}     & \zqh{-}    & \zqh{74.91}    & \zqh{-}   & \zqh{-}     \\
		& \zqh{Dual-MRC~\cite{mao2021joint}} & \zqh{-}     & \zqh{75.97}    & \zqh{-}       & \zqh{82.04}       & \zqh{-}   & \zqh{-}     & \zqh{-}    & \zqh{73.59}    & \zqh{-}   & \zqh{-}     \\
		& \zqh{DCRAN-BERT~\cite{oh2021deep}} & \zqh{-}     & \zqh{77.02}    & \zqh{-}       & \zqh{80.64}       & \zqh{-}   & \zqh{-}     & \zqh{-}    & \zqh{76.30}    & \zqh{-}   & \zqh{-}     \\
		& \zqh{DGEDT-BERT \cite{tang2020dependency}}    & \zqh{79.80 }   & \zqh{75.60}    & \zqh{86.30}     & \zqh{80.00}       & \zqh{77.90}     & \zqh{75.40}     & \zqh{84.00}           & \zqh{71.00}          & \zqh{91.90}   & \zqh{79.00}      \\
		& \zqh{T-GCN-BERT~\cite{tian2021aspect}}     & \zqh{80.88}     & \zqh{77.03}     & \zqh{86.16}         & \zqh{79.95}     & \zqh{76.45}    & \zqh{75.25}     & \zqh{85.26}           & \zqh{71.69}          & \zqh{92.32}   & \zqh{77.29}     \\
		& \zqh{DM-GCN-BERT \cite{pang-etal-2021-dynamic}}    & \zqh{80.22}  & \zqh{77.28}     & \zqh{87.66}    & \zqh{82.79}    & \zqh{78.06}    & \zqh{77.36}      & \zqh{-}      & \zqh{-}   & \zqh{-}       & \zqh{-}        \\
		& \zqh{Sentic GCN-BERT \cite{liang2022aspect}}     & \zqh{82.12}    & \zqh{79.05}    & \zqh{86.92}     & \zqh{81.03}      & \zqh{-}     & \zqh{-}     & \zqh{85.32}           & \zqh{71.28}          & \zqh{91.97}   & \zqh{79.56}      \\
		& \zqh{SSEGCN-BERT \cite{zhang2022ssegcn}}     & \zqh{81.01}    & \zqh{77.96}    & \zqh{87.31}     & \zqh{81.09}      & \zqh{77.40}     & \zqh{76.02}     & \zqh{-}           & \zqh{-}          & \zqh{-}   & \zqh{-}      \\
		&DCRAN~\cite{oh2021deep} & \zqh{-}    & \zqh{77.02}    & \zqh{-}     & \zqh{78.67}      & \zqh{-}     & \zqh{-}     & \zqh{-}           & \zqh{73.30}          & \zqh{-}   & \zqh{-}      \\
		& \zqh{dotGCN-BERT \cite{chen2022discrete}}     & \zqh{81.03}    & \zqh{78.10}    & \zqh{86.16}     & \zqh{80.49}      & \zqh{78.11}     & \zqh{77.00}     & \zqh{85.24}           & \zqh{72.74}          & \zqh{\textbf{93.18}}   & \zqh{\underline{82.32}}      \\
		\cmidrule{2-12}
		& \zqh{RGAT-BERT \cite{bai2021rgat}}  & \zqh{80.94}    & \zqh{78.20}     & \zqh{86.68}     & \zqh{80.92}     & \zqh{76.28}   & \zqh{75.25}      & \zqh{83.64$^{\star}$}           & \zqh{66.18$^{\star}$}          & \zqh{90.16$^{\star}$}   & \zqh{71.13$^{\star}$}   \\
		& \multirow{1}{*}{\zqh{\textbf{RGAT-BERT+UIKA (Ours)}}} &\zqh{82.03} &\zqh{78.83} &\zqh{87.25} &\zqh{81.98} &\zqh{78.55} &\zqh{77.84} &\zqh{86.40} &\zqh{68.11} &\zqh{91.87} &\zqh{75.28} \\		
		\cdashline{2-12}
		& \zqh{DualGCN-BERT \cite{li-etal-2021-dual-graph}}     & \zqh{81.80}     & \zqh{78.10}    & \zqh{87.13}   & \zqh{81.16}   & \zqh{77.40}      & \zqh{76.02}      & 84.25$^{\star}$     & 69.54$^{\star}$   & 89.22$^{\star}$       & 72.40$^{\star}$                    \\
		& \multirow{1}{*}{\textbf{DualGCN-BERT+UIKA (Ours)}}     & 82.43    & 78.71   & \underline{87.90} & 81.97    & 77.86    &76.75    & \underline{86.82}     & 69.80   & 90.81      & 73.13   \\		
		\cdashline{2-12}
		& \zqh{KGAN-BERT~\cite{zhong2023KGAN}}   & \zqh{\underline{82.66}}    & \zqh{\underline{78.98}}     & \zqh{87.15} & \zqh{\underline{82.05}} & \zqh{\underline{79.97}}   & \zqh{\underline{79.39}}    & \zqh{86.21}   & \zqh{\underline{74.20}}          & \zqh{92.34}   & \zqh{81.31}     \\
		& \multirow{1}{*}{\zqh{\textbf{KGAN-BERT+UIKA (Ours)}}} &\zqh{\textbf{83.21}}  &\zqh{\textbf{79.57}} &\zqh{\textbf{87.92}} &\zqh{\textbf{82.82}} &\zqh{\textbf{80.33}} &\zqh{\textbf{80.21}} &\zqh{\textbf{87.43}} &\zqh{\textbf{75.12}}  &\zqh{\underline{92.89}} &\zqh{\textbf{82.43}} \\		
        \Xhline{1.2pt}
	\end{tabular}
\end{table*}

\subsection{UIKA on Different Architectures}
\label{section3.3}
Our proposed UIKA framework is effective for all DNN-based methods (i.e., CNN-, LSTM- and syntax-based methods) on the ABSA task. In this section, we describe how we implement the UIKA framework for all types of methods to promisingly improve their performance on ABSA. Specifically, we choose GCAE \cite{xue2018aspect}, ATAE-LSTM \cite{wang2016attention} and ASGCN \cite{zhang-etal-2019-aspect} to represent the CNN-, LSTM- and syntax-based methods, respectively\footnote{Notably, we state that our framework works well on various cutting-edge ABSA models, \textit{e.g.,} R-GAT~\cite{wang2020relational} and DM-GCN~\cite{pang-etal-2021-dynamic}. However, we use the simple models, \textit{i.e.}, GCAE, ATAE-LSTM and ASGCN, owing to their typicality and applicability}. For ease of illustration, we only describe the implementation of the UIKA framework on the syntax-based ASGCN in detail, while the others are similar to ASGCN. 

In practice, we first construct the sampled $SD_R'$ via the proposed coarse-to-fine retrieval strategy. Then, we build a pretraining model, denoted as ASGCN-pre, which is the same as ASGCN except that the former does not apply the position component since $SD_R'$ does not provide reliable position information. The ASGCN-pre is used to train on $SD_R'$ and obtain the pretrained model $M_1$. 
According to the second stage, we make the knowledge guidance and the learner models share the same architecture with the ASGCN-pre and train them according to the described procedures. In particular, while initializing the knowledge guidance and learner models, we ignore the parameters of the fully connected layer since $TD$ has a different number of label categories with $SD_R'$. Specifically, $TD$ contains three sentiment polarities (positive, negative and neutral), while $SD_R'$ usually contains two types (positive and negative). After the second stage, we regard the obtained learner model as the output pretrained model $M_2$. In the third stage, we initialize the pure ASGCN with the weights from $M_2$. Finally, we train the ASGCN on $TD$ to further finetune those weights and obtain the final classification model $M_3$.

\section{Experiments}
\label{section4}
\subsection{Datasets and Experimental Settings}
\zqh{Experiments are conducted on six public standard aspect-level datasets, \textit{i.e., }, Laptop14, Restaurant14, Twitter~\cite{dong2014adaptive}, Restaurant15, Restaurant16 and MAMS~\cite{jiang2019challenge}. The Laptop14 and Restaurant14 datasets are from the SemEval2014 ABSA challenge \cite{pontiki-etal-2014-semeval}, and Restaurant15 and Restaurant16 are from the SemEval2015~\cite{pontiki2015semeval} and SemEval2016~\cite{pontiki2016semeval} challenges, respectively.}
Following Tang \emph{et al.} \cite{tang2019progressive}, we remove a few instances with conflicting sentiment polarity and then list the statistics of these datasets in Tab.~\ref{table0}.
Additionally, two sentence-level Amazon and Yelp review datasets \cite{zhang2015nips} are adopted as pretraining datasets. Notably, we set N and K of Algorithm~\ref{Algorithm 1} as 500 and 300\footnote{\zqh{The detailed analysis of both hyperparameters can be found in Sec~\ref{sec_ablation}. }}, respectively. In other words, for both pretraining datasets, we employ our sampling strategy to select the top 500 coarse-grained instances and top 300 fine-grained instances for each sentence of the target datasets.

In our implementation, we basically utilize GloVe vectors \cite{pennington2014glove} to initialize the word embeddings. We also carefully validate the effectiveness of our method based on a powerful pretrained language model, \textit{i.e., }, BERT~\cite{devlin2018bert}\footnote{\zqh{Although GPT2~\cite{radford2019language} is also a powerful PLM, it is usually used to tackle language generation tasks and may be less helpful for ABSA tasks. Thus, we choose the widely used BERT as the PLM.}}.
\zqh{The learning rates are empirically set as 1e-3 for the GloVe-based models and 2e-5 for the BERT-based models.}
The batch sizes are 256, 64 and 64 for the first, second and third stages, respectively. To avoid overfitting, we apply dropout on the embedding layer and the final sentence representation with a dropout rate of 0.2. We adopt Adam~\cite{kingma2015adam} to fulfil the optimization and training for the models.

\zqh{To investigate the effectiveness of our UIKA, we mainly apply it to improve three cutting-edge baseline models, \textit{i.e.}, RGAT~\cite{bai2021rgat}, DualGCN~\cite{li-etal-2021-dual-graph}} and KGAN~\cite{zhong2023KGAN}. For comparison, we also report the performance of other state-of-the-art methods and the most typical ones in this field. In particular, we include the following methods: 
\zqh{\textbf{(a) GloVe-based}: TD-LSTM~\cite{tang2016effective}, IAN~\cite{ma2017interactive},\zqh{ATAE-LSTM~\cite{wang2016attention}}, \zqh{GCAE~\cite{xue2018aspect}}, \zqh{RAM~\cite{chen2017recurrent}}, TransCap~\cite{chen2019transfer}, TNet-AS \cite{li2018transformation}, MGAN~\cite{li2019exploiting}, TNet-ATT~\cite{tang2019progressive}, MCRF-SA~\cite{xu-etal-2020-aspect}, \zqh{ASCNN~\cite{zhang-etal-2019-aspect}}, \zqh{RACL~\cite{chen2020relation}}, DGEDT~\cite{tang2020dependency}, R-GAT~\cite{wang2020relational}, DM-GCN~\cite{pang-etal-2021-dynamic}, Sentic-LSTM~\cite{ma2018sentic}, Sentic GCN~\cite{liang2022aspect}, SSEGCN~\cite{zhang2022ssegcn}.
\textbf{(b) BERT-based}: Vanilla BERT~\cite{devlin2018bert}, R-GAT-BERT~\cite{wang2020relational}, RACL-BERT~\cite{chen2020relation}, Dual-MRC~\cite{mao2021joint}, DCRAN-BERT~\cite{oh2021deep}, DGEDT-BERT~\cite{tang2020dependency}, T-GCN-BERT~\cite{tian2021aspect}, DM-GCN-BERT~\cite{pang-etal-2021-dynamic}, Sentic GCN-BERT~\cite{liang2022aspect}, SSEGCN-BERT~\cite{zhang2022ssegcn}, DCRAN~\cite{oh2021deep} and dotGCN~\cite{chen2022discrete}.
}

Notably, in our experiments, we report the averaged results over 5 random seeds to avoid stochasticity. Following the previous work~\cite{zhou2020sk}, we also apply a student t-test~\cite{mendenhall2012introduction} to conduct the statistical analysis.

\begin{table}[tp]
	\centering
	\caption{\zqh{Contrastive performance on MASA~\cite{jiang2019challenge}.}}
	\label{table2_2}
	\begin{tabular}{clcc}
		\Xhline{0.8pt}
		\multirow{2}{*}{\zqh{\textbf{Embedding}}}&\multirow{2}{*}{\zqh{\textbf{Method}}} & \multicolumn{2}{c}{\zqh{\textbf{MAMS}}}             \\ \cline{3-4} 
		&& \zqh{Acc. (\%)}         & \zqh{F1. (\%)}           \\  \hline \hline
		\multirow{7}{*}{\zqh{\textbf{GloVe}}} 
		&\zqh{IAN~\cite{ma2017interactive}}         & \zqh{76.60 }     & \zqh{-}        \\
		&\zqh{AOA~\cite{huang2018aspect}}        &\zqh{77.26}   &  \zqh{-}     \\
		&\zqh{GCAE~\cite{xue2018aspect}}        & \zqh{77.59}   &\zqh{ -}       \\  
		&\zqh{CapsNet~\cite{jiang2019challenge}}        & \zqh{79.78}   & \zqh{-}       \\ 
		&\zqh{CDT~\cite{sun2019aspect}}        & \zqh{80.80}   & \zqh{79.79}      \\
		\cdashline{2-4} 
		&\zqh{RGAT~\cite{bai2021rgat}}        & \zqh{\underline{81.75}}  & \zqh{\underline{80.87}}      \\ 
		&\zqh{RGAT+UIKA (Ours)}      &\zqh{\textbf{83.21}}   & \zqh{\textbf{81.52}}      \\ 
		\midrule
		\multirow{6}{*}{\zqh{\textbf{BERT}}} 
		&\zqh{BERT-SPC~\cite{devlin2018bert}}          & \zqh{82.82}      & \zqh{81.90}        \\
		&\zqh{CapsNet-BERT~\cite{jiang2019challenge}}       &\zqh{83.39}   &  \zqh{-}     \\
		&\zqh{T-GCN-BERT~\cite{tian2021aspect}} &\zqh{83.38} &\zqh{82.77} \\
		&\zqh{dotGCN~\cite{chen2022discrete}} &\zqh{\underline{84.95}} &\zqh{\textbf{84.44}} \\
		\cdashline{2-4}
	    &\zqh{RGAT-BERT~\cite{bai2021rgat}}       & \zqh{84.52}     & \zqh{83.74}      \\ 
		&\zqh{RGAT-BERT+UIKA (Ours)}        &  \zqh{\textbf{85.43}}    & \zqh{\underline{84.28}}      \\
		\Xhline{0.8pt}
	\end{tabular}
\end{table}

\subsection{Main Results and Analysis}
\begin{figure}[tp]
	\centering
	\includegraphics[width=0.48\textwidth]{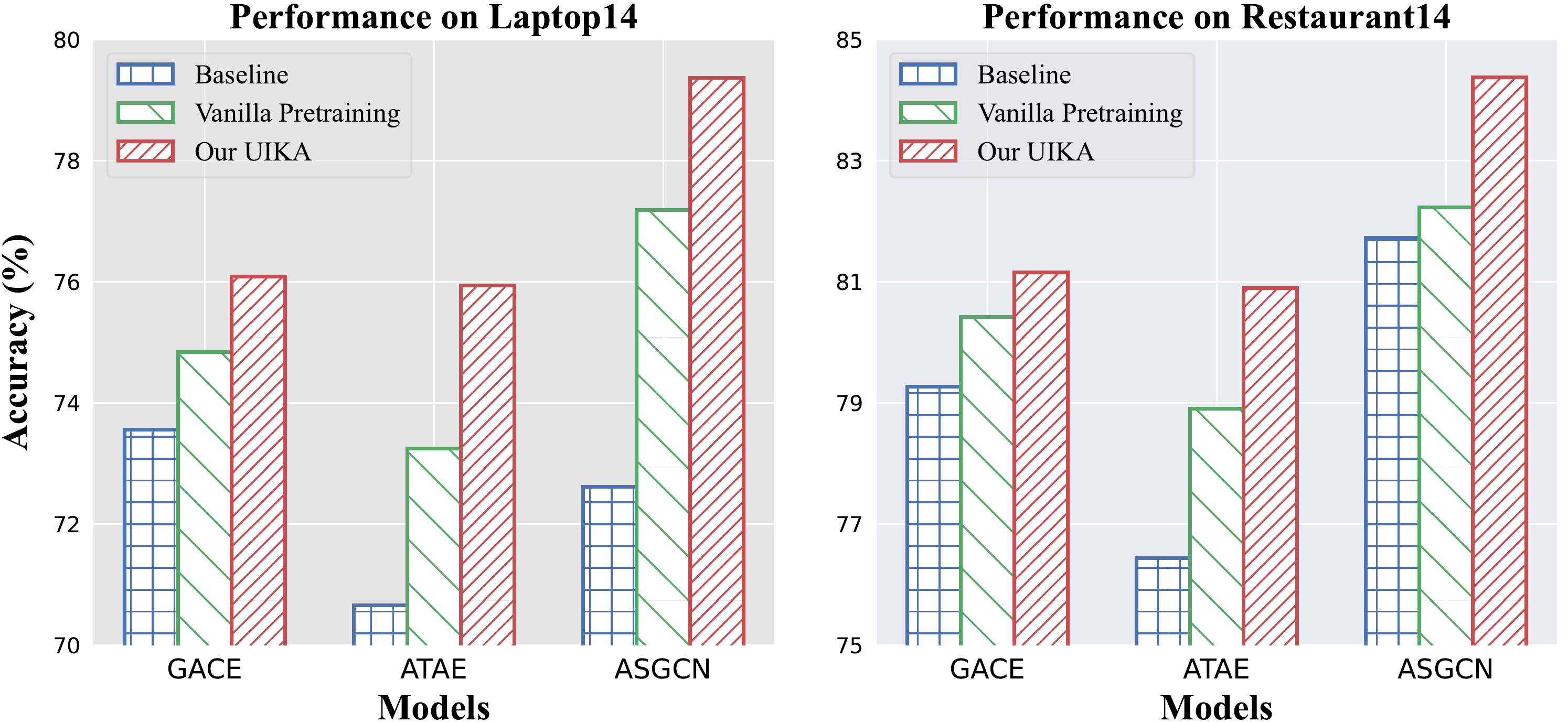} 
	\caption{\zqh{Analysis of whether our UIKA can be applied in various model scenarios. We compare our UIKA with the vanilla pretraining method. }}
	\label{fig5}
\end{figure}

\zqh{Experimental results are presented in Tab.~\ref{table1}. Note that we use Amazon as the pretraining dataset. The analysis of different pretraining datasets can be found in Sec.~\ref{sec_pd}. From the results of Tab.~\ref{table1}, we find the following:

(i) As seen, in the GloVe-based setting, with the help of our UIKA framework, the baseline ``KGAN'' outperforms all the other methods and achieves the best performance on most ABSA benchmarks. Specifically, UIKA consistently produces a performance improvement for all the benchmarks, where ``KGAN+UIKA'' gains up to 1.07\% score. These results prove the effectiveness of our UIKA. In addition to KGAN, we also apply our UIKA method to RGAT~\cite{bai2021rgat}, which is also a cutting-edge ABSA baseline. As seen, UIKA is also beneficial to RGAT, as it consistently and effectively improves the performance of RGAT. Encouragingly, with the help of UIKA, RGAT outperforms most of its powerful counterparts, \textit{e.g., }, DualGCN~\cite{pang-etal-2021-dynamic} and DM-GCN~\cite{li-etal-2021-dual-graph}. These results indicate that UIKA works extremely well on various cutting-edge ABSA models.

(ii) Furthermore, we evaluate our framework on the BERT-based settings. Compared to the other BERT-based models, our improved RGAT-BERT and KGAN-BERT achieve comparable and even better performances on most ABSA benchmarks. Specifically, trained with UIKA, KGAN-BERT achieves new record-breaking state-of-the-art performances, continuing to prove the effectiveness of our framework. We conclude that UIKA is complementary to powerful pretrained language models, such as BERT.

(iii) We additionally conduct experiments on the MAMS dataset and report the results in Tab.~\ref{table2_2}. Notably, since many models were not evaluated on MAMS, and it is difficult for us to reproduce all the methods, we only report some results of the most often compared models. As seen, our UIKA boosts the performance of RGAT on this difficult benchmark. Specifically, the accuracy performance gains of RGAT and RGAT-BERT are up to 1.46\% and 0.91\%, respectively.

(iv) As stated in Sec.~\ref{section3.3}, our proposed UIKA framework is effective for all the DNN-based methods on the ABSA task. Here, to verify it, we conduct experiments on several typical baseline models, \textit{i.e.}, GCAE~\cite{xue2018aspect} (CNN-based), ATAE-LSTM~\cite{wang2016attention} (LSTM-based) and ASGCN~\cite{zhang-etal-2019-aspect} (Syntax-based). Fig.~\ref{fig5} illustrates the results. For reference, we also report the results of vanilla pretraining strategies. As seen, our UIKA consistently and significantly improves the performance among all the model types on both datasets. These results demonstrate its universality.

(v) Some readers may wonder whether our UIKA is superior to other task-specific pretraining works~\cite{xu2019bert,li2021learning,you2022ask,yong2023sgpt} in the ABSA field. Here, given the same pretraining data, we compare our UIKA with its counterparts in detail\footnote{Notably, for the hyper-parameter settings of other methods, we basically follow their original papers for better performance.}. 
Considering that most of these compared methods are based on BERT, we use RGAT-BERT as the baseline. Notably, to analyse the efficiency of UIKA, we also report the extra pretraining budgets for each method. Tab.~\ref{table2_3} lists the detailed results, which show that UIKA achieves the most performance gains. Moreover, different from the counterparts that need to continue pretraining the BERT using the MLM objective~\cite{devlin2018bert}, UIKA has less extra computational overhead. These results prove the superiority and efficiency of our framework.
}




\begin{table}[tp]
	\centering
	\caption{\zqh{Comparison of different pretraining strategies on Laptop14 and Restaurant14.}}
	\label{table2_3}
	\begin{tabular}{lccccc}
		\Xhline{0.8pt}
		\multirow{2}{*}{\textbf{\zqh{Method}}}&\multirow{1}{*}{\textbf{\zqh{Budget}}} & \multicolumn{2}{c}{\textbf{\zqh{Laptop14}}}    & \multicolumn{2}{c}{\textbf{\zqh{Restaurant14}}}          \\ \cline{2-6} 
		&\zqh{training time}& \zqh{Acc.}         & \zqh{F1}          & \zqh{Acc.}         & \zqh{F1}                \\  \hline \hline
		\zqh{RGAT-BERT~\cite{bai2021rgat}}  &\zqh{-} & \zqh{80.94}    & \zqh{78.20}     & \zqh{86.68}     & \zqh{80.92}       \\ 
		\midrule
		\multicolumn{6}{l}{\zqh{\textit{Equipped with task-specific pretraining}}} \\ \hdashline
		\quad \zqh{-w/ vanilla Pretrain}    & \zqh{0.50 h} &\zqh{81.41}	&\zqh{77.77}  &   \zqh{86.07}	&\zqh{79.43}      \\
		\quad \zqh{-w/ BERT PT~\cite{xu2019bert}} &\zqh{3.21 h} & \zqh{82.01}    & \zqh{\textbf{79.31}}   &  \zqh{86.27}	&\zqh{79.44}      \\
		\quad \zqh{-w/ SCAPT ~\cite{li2021learning}}   & \zqh{3.86 h} &\zqh{81.25} &\zqh{78.21} &\zqh{86.92} &\zqh{80.35}       \\ 
		\quad \zqh{-w/ Ask-BERT~\cite{you2022ask}} &\zqh{1.23 h} &\zqh{81.02} &\zqh{78.30} &\zqh{86.90} &\zqh{80.78} \\
		\quad \zqh{-w/ SGPT~\cite{yong2023sgpt}} &\zqh{3.53 h} &\zqh{80.98} &\zqh{78.52} &\zqh{86.56} &\zqh{80.43}  \\
		\quad \zqh{-w/ UIKA (Ours)}  &\zqh{0.67 h} &\zqh{\textbf{82.03}} &\zqh{78.83} &\zqh{\textbf{87.25}} &\zqh{\textbf{81.98}}      \\ 
		\Xhline{0.8pt}
	\end{tabular}
\end{table}

\begin{figure}[tp]
	\centering
	\includegraphics[width=0.48\textwidth]{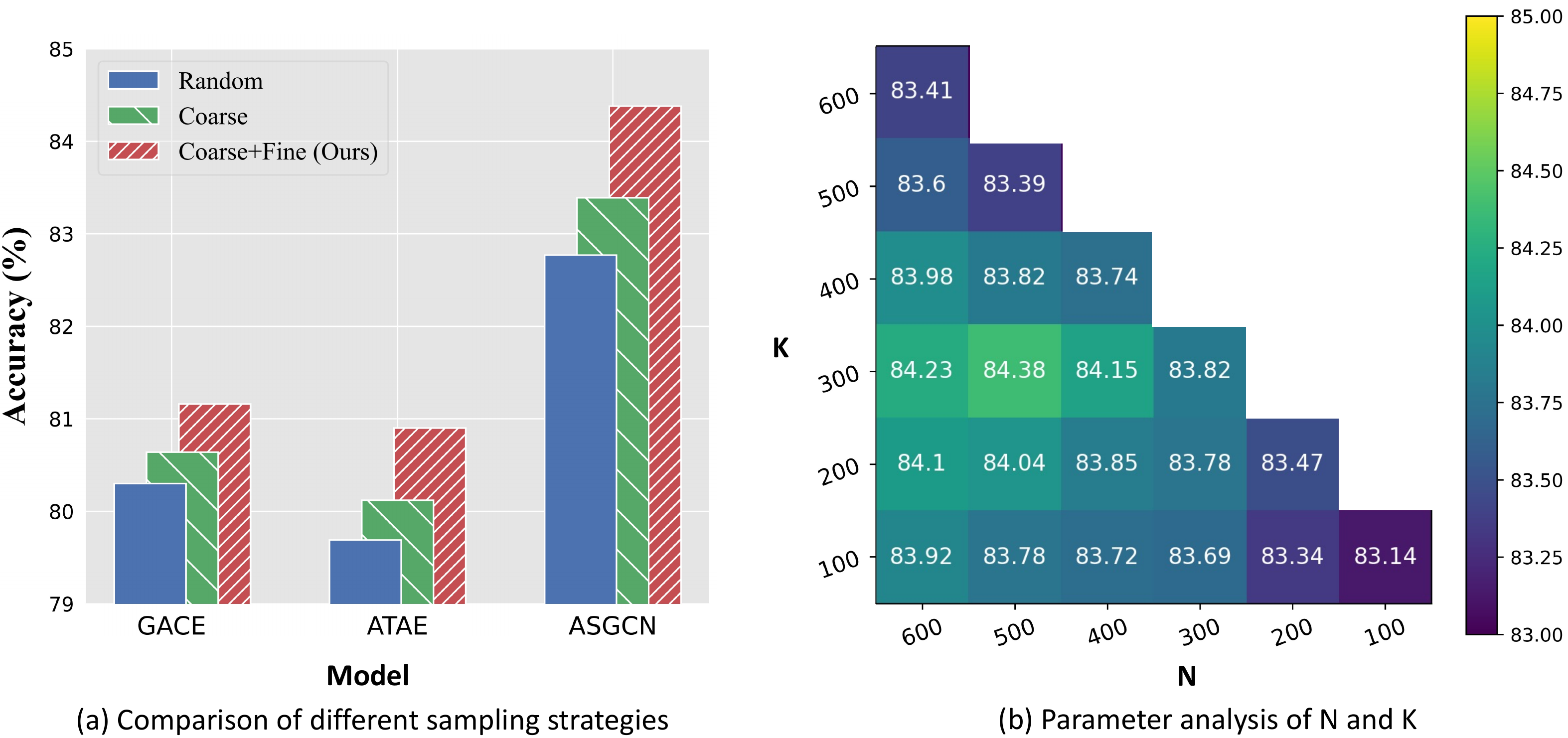} 
	\caption{\zqh{\textbf{(a) Comparison of different sampling strategies}. `Random" and ``Coarse" denote a random sampling strategy and coarse-grained sampling strategy, respectively (\textit{i.e.} directly using BM25). \textbf{(b) Parameter analysis of N and K}. Note that we use the ASGCN as the baseline and evaluate its performance on Restaurant14.}}
	\label{fig6}
\end{figure}

\subsection{Ablation Study}
\label{sec_ablation}
We conduct extensive ablation studies to further investigate the effect of the proposed sampling strategy and the important components in our UIKA framework. \zqh{Notably, due to limited computational resources, we mainly use the simpler and low-cost models (GCAE/ATAE/ASGCN) as the baselines in the following experiments.}

\subsubsection{Effect of the proposed sampling strategy}
As stated in Sec.~\ref{section1}, we introduced a novel coarse-to-fine sampling strategy to align the instances between pretraining and the target ABSA datasets. To prove the effectiveness of the sampling strategy, we compare the proposed strategy (denoted as \textbf{Ours)} with the other strategies: 1) \textbf{Random}: the instances are randomly selected from the pretraining dataset; 2) \textbf{Coarse}: a coarse-grained strategy (\textit{i.e.} BM25) is used to select the related instances. For a fair comparison, we sample the same number of instances from the Amazon dataset for each strategy.

The results in Fig.~\ref{fig6} (a) show that all the baseline models upon introduction to our strategy consistently outperform the other strategies on both of the evaluated datasets, indicating the effectiveness of the introduced strategy. More specifically, for the ASGCN-UIKA evaluated on Laptop14, pretraining with our sampling strategy (\textit{Acc: 79.37\%, F1: 76.05\%}) outperforms the random-based strategy (\textit{Acc: 77.66(+1.71)\%, F1: 73.74(+2.31)\%}) by a large margin. We attribute such performance gains to the narrowed distribution gap between the pretraining and target ABSA datasets by introducing instance-level alignments.

\zqh{Moreover, as mentioned in Sec.~\ref{section3}, we use two important hyperparameters (\textit{N} and \textit{K}) to control the number of coarse-to-fine sampling sentences. Here, to investigate the effect of both weights, we evaluate the ASGCN-UIKA on the Restaurant14 task with different combinations of \textit{N} and \textit{K}. Fig.~\ref{fig6} (b) illustrates the detailed results. Generally, pretraining with more target-related samples can achieve better performances. In the case of \textit{N}=500 and \textit{K}=300, UIKA achieves the best performance, thus it is the default setting.
}

\begin{table}[]
	\centering
	\caption{Experimental results of GCAE on Laptop14 and Restaurant14. \textbf{S.} and \textbf{F.} respectively, denote the first and third stages in the UIKA framework, while \textbf{R.} and \textbf{E.} are the two components of the second stage. Note that we also employ the sampled Amazon dataset for pretraining in this study.  The best results are in bold.}
	\label{table3}
	\begin{tabular}{cccccccc}
		\Xhline{0.8pt}
		\multirow{2}{*}{\textbf{S.}} &\multirow{2}{*}{\textbf{R.}} &\multirow{2}{*}{\textbf{E.}} &\multirow{2}{*}{\textbf{F.}} & \multicolumn{2}{c}{\textbf{Laptop14}}    & \multicolumn{2}{c}{\textbf{Restaurant14}}          \\ \cline{5-8} 
		&&&& Acc.(\%)         & F1.(\%)          & Acc.(\%)        & F1.(\%)                 \\  \hline \hline
		&&&\checkmark          & 73.56     & 67.84       & 79.27      & 67.66       \\
		\checkmark&&&\checkmark          &  74.22    & 68.58       &  79.79     &  69.65      \\
		\checkmark&\checkmark && \checkmark            &74.69      & 69.56       & 80.20      & 70.71   \\
		\checkmark&& \checkmark&  \checkmark       & 74.53     & 70.04       & 79.95      & 70.48    \\
		\checkmark&\checkmark& \checkmark&            & 70.62     & 61.82       &  78.12     & 63.51   \\
		\checkmark&\checkmark&\checkmark&\checkmark             & \textbf{76.09} & \textbf{70.66} & \textbf{81.16}    & \textbf{71.56}  \\
		\Xhline{0.8pt}
	\end{tabular}
\end{table}

\begin{table}[tp]
	\centering
	\caption{Parameter analysis of the weight factor $\alpha$ in Eq.~\ref{eq2}. Notably, ``W/o $\alpha$" means not using $\alpha$ (\textit{i.e.} $\alpha=0$), ``Constant" refers to the constant $\alpha$ and ``Adaptive (ours)'' denotes the annealing function introduced in our framework. The symbol $\downarrow$ denotes a drop of performance compared to ``Adaptive (ours)''.}
	\label{table4}
	\begin{tabular}{ccccc}
		\Xhline{0.8pt}
		\multirow{2}{*}{\textbf{Approach}} & \multicolumn{2}{c}{\textbf{Laptop14}}    & \multicolumn{2}{c}{\textbf{Restaurant14}}          \\ \cline{2-5} 
		&Acc.(\%)         & F1.(\%)          & Acc.(\%)         & F1.(\%)                \\  \hline \hline
		W/o $\alpha$       & $\downarrow$1.56 & $\downarrow$0.62 & $\downarrow$1.21    & $\downarrow$1.08    \\
		Constant       & $\downarrow$0.46  & $\downarrow$0.04  & $\downarrow$0.16    & $\downarrow$0.38            \\ 
		Adaptive (ours)          &76.09 &70.66 &81.16   &71.56    \\
		\Xhline{0.8pt}
	\end{tabular}
\end{table}

\subsubsection{Effect of Multiple Important Components in UIKA}
To investigate the impact of the essential components in our proposed UIKA, we report the results of different component combinations in Tab.~\ref{table3}. Notably, \textbf{S.} presents the first initial pretraining stage; \textbf{R.} and \textbf{E.}, respectively denote the representation consistency loss function $\mathcal{L}_{r}$ in Eq.~\ref{eq2} and the exponential moving average method in Eq.~\ref{eq3} of the second stage; \textbf{F.} refers to the final finetuning stage. Specifically, in the case of training without the exponential moving average method (\textit{i.e.} w/o \textbf{E.}), we only optimize the knowledge guidance model via the guidance loss $\mathcal{L}_{G}$ and do not update the parameters of the learner model. Correspondingly, the final classifier model is initialized with the parameters of the knowledge guidance model and then finetuned on the target dataset. For ease of illustration, we only report the results of baseline GCAE on Laptop14 and Restaurant14. Note that the results on the other baselines are similar.

It is obvious that the model with all the components performs best,
and all the components in the different stages are consistently beneficial to our UIKA. Specifically, the \textbf{R.} can increase the macro-F1 score of Restaurant14 from 69.07\% to 70.71\%, and the \textbf{E.} improves the macro-F1 score over 1.46\% on Laptop14. These results demonstrate that both components used in our framework are effective and practicable. Additionally, the model trained without the final finetuning stage expectedly performs poorly since the combination of the first two stages attempts to learn the domain-invariant knowledge but is not a good classifier on the target dataset. \zqh{In the third stage, we use the supervision of the target dataset to further fine-tune the parameters (especially the classifier head) of the learner model, thus making it learn more target-side information.} This indicates the importance of finetuning the pretrained model on the target dataset.

\begin{table}[tp]
	\centering
	\caption{Parameter analysis of the weight factor $\beta$. The results with GCAE-UIKA pretraining with the Amazon dataset. The best results are in bold.}
	\label{table5}
	\begin{tabular}{ccccc}
		\Xhline{0.8pt}
		\multirow{2}{*}{\textbf{Values of $\beta$}} & \multicolumn{2}{c}{\textbf{Laptop14}}    & \multicolumn{2}{c}{\textbf{Restaurant14}}          \\ \cline{2-5} 
		&Acc.(\%)         & F1.(\%)          & Acc.(\%)         & F1.(\%)                 \\  \hline \hline
		0.3         &73.84       &68.72          & 79.60          & 69.27      \\
		0.5       &74.53 &69.59 & 79.86    & 69.96    \\
		0.7       &74.69  &69.25 & 79.95    & 70.98            \\ 
		0.9       &75.16 & 69.98 & 80.38    & 71.03                   \\
		0.99       &\textbf{76.09} &\textbf{70.66} &\textbf{81.16}    &\textbf{71.56 }                  \\
		0.999       & 75.00 &69.97 &80.56    & 70.07                   \\
		\Xhline{0.8pt}
	\end{tabular}
\end{table}

\begin{table*}[tp]
	\centering
	\caption{The words enclosed in [] are aspect terms, and \textit{p}, \textit{n}, \textit{o} denote the true polarities. P, N and O respectively denote the positive, negative and neutral predictions. The symbols \ding{51} and \ding{55} indicate the correct and wrong predictions, respectively.}
	\label{table6}
	\begin{tabular}{m{7cm}|ccccc}
		\Xhline{0.8pt}
		Sentences                                                                                                    &TNet-AS &TransCap     &TNet-ATT  &ASGCN &ASGCN-UIKA \\ \hline
		1. Great $[food]_p$ but the $[service]_n$ is dreadful.      
		&(P\ding{51}, N\ding{51})   &(P\ding{51}, N\ding{51})      &(P\ding{51}, N\ding{51})      &(P\ding{51}, N\ding{51})         &(P\ding{51}, N\ding{51})         \\  \hline
		2. The $[staff]_n$ should be a bit more friendly.         &P\ding{55}    &N\ding{51}   &P\ding{55}       &P\ding{55}     &N\ding{51}   \\   \hline
		3. The $[food]_o$ did take a few extra minutes to come, but cute $[waiters]_p$' jokes and friendliness made up for it.  &(P\ding{55}, P\ding{51})    &(P\ding{55}, P\ding{51})    &(O\ding{51}, P\ding{51})   &(P\ding{55}, P\ding{51})       &(O\ding{51}, P\ding{51})    \\ \hline
		4. The $[folding$ $chair]_n$ I was seated at was uncomfortable.       
		&O\ding{55}            &N\ding{51}    &N\ding{51}    &O\ding{55}        &N\ding{51}        \\  
		\Xhline{0.8pt}
	\end{tabular}
	
\end{table*}

\subsection{Analysis of Hyper-Parameters}
\label{section4.4}
\subsubsection{Analysis of $\alpha$}
In $\mathcal{L}_{G}$ of Eq.~\ref{eq2}, we introduced a linear annealing function to dynamically adjust the value of $\alpha$. To validate the effect of this function, we conduct the contrastive experiments as follows: 1) ``W/o $\alpha$'': we remove the $\alpha$, \textit{i.e.} $\mathcal{L}_{G} = \mathcal{L}_{c}$; 2) ``Constant'': we empirically set $\alpha$ as the constant, \textit{i.e.} $\alpha=0.7$; 3) ``Adaptive (ours)'': training with our introduced annealing function. The results of GCAE with the different $\alpha$ are shown in Tab.~\ref{table4}.

We see that the introduced annealing function can boost the performance promisingly, compared to the other approaches (\textit{i.e.} without or with constant $\alpha$), which validates the effectiveness of this function. Specifically, we find that the dynamic $\alpha$ could aid the knowledge guidance model in aggressively learning the target domain knowledge at the beginning of the second pretraining stage while learning more conservatively at the end. In this way, the knowledge guidance model not only learns the target knowledge but also avoids the problem of catastrophic forgetting~\cite{rusu2016progressive}.

\subsubsection{Analysis of $\beta$}
The $\beta$ in Eq.~\ref{eq3} is an important parameter that controls the degree of knowledge transferred from the knowledge guidance model to the learner model. In this subsection, we analyse the effect of $\beta$ on the Laptop14 and Restaurant14 datasets. Specifically, we vary $\beta$ in the range [0,1] and evaluate the performance of the GCAE baseline trained with the proposed UIKA framework. Tab.~\ref{table5} shows the comparative results.

From the results, we observe the following: (1) When $\beta$ is too large, \textit{e.g.} 0.999, it is difficult for the UIKA framework to transfer knowledge from the knowledge guidance model to the learner model, thus hindering the learning of domain-invariant knowledge. (2) On the other hand, if $\beta$ is too small, \textit{e.g.} 0.3 and 0.5, the update speed of the learner model is too large, and the prior knowledge may suffer from catastrophic forgetting. (3) In the case of $\beta$=0.99, the UIKA framework achieves the highest performance improvement for both datasets, showing that 0.99 is a good trade-off between learning speed and performance. We thereby leave $\beta=0.99$ as the default setting in the main experiments.

\begin{table*}[tp]
	\centering
	\caption{Some reviews and their sentence-level labels of Amazon and Yelp datasets. Notably, the words in italics denote the candidate aspect terms, while the ones in bold are the final extracted aspect terms.}
	\label{table7}
	\begin{tabular}{m{5.5cm}c|m{5.5cm}c}
		\Xhline{0.8pt}
		Amazon reviews  &Sentence-level label & Yelp reviews &Sentence-level label  \\ \hline \hline
		
		This \emph{\textbf{grill}} is one of its kind...It can be used anywhere without so much hard work..it fits snugly anywhere....     & Positive    &  Overpriced, salty and overrated!!! Why this \emph{\textbf{place}} is so popular I will never understand.   &Negative   \\  \hline
		
		These \emph{\textbf{pads}} are great. Took me about a week to train my puppy to pee and poop on them. Now no "surprises" around. Very satisfied and I would recommend them to anyone.  &Positive   & Enjoyed it through and through. Had the big unit \emph{\textbf{hot dog}}. It was actually delicious. Will go back whenever im in Arizona.    &Positive          \\	\hline
		
		The \emph{phone} is great. But the \emph{\textbf{battery life}} is terrible. Also there is no battery life indicator to let you know when its low. ... &Negative  &  Friendly \emph{staff} and nice selection of vegetarian options.  \emph{\textbf{Food}} is just okay, not great.  Makes me wonder why everyone likes Food Fight so much.   & Negative       \\  
		\Xhline{0.8pt}
	\end{tabular}
\end{table*}

\subsection{Case Study}
In this subsection, to have a closer look, we select some instances for extensive case studies.
We first present some examples in Tab.~\ref{table6} to show the improvement of the UIKA framework on the baseline ASGCN.
In this case, aspect terms are enclosed in [], and the subscripts $p$, $n$ and ${o}$ denote the true polarity ``positive'', ``negative'' and ``neutral'', respectively. 
Specifically, we observe that the pure ASGCN performs worse than TNet-AS, TransCap and TNet-ATT. However, with the help of UIKA, the ASGCN-UIKA could achieve a better performance and make correct predictions for all the sentences, indicating the effectiveness of our UIKA framework.

\begin{figure}[tp]
	\centering
	\includegraphics[width=0.45\textwidth]{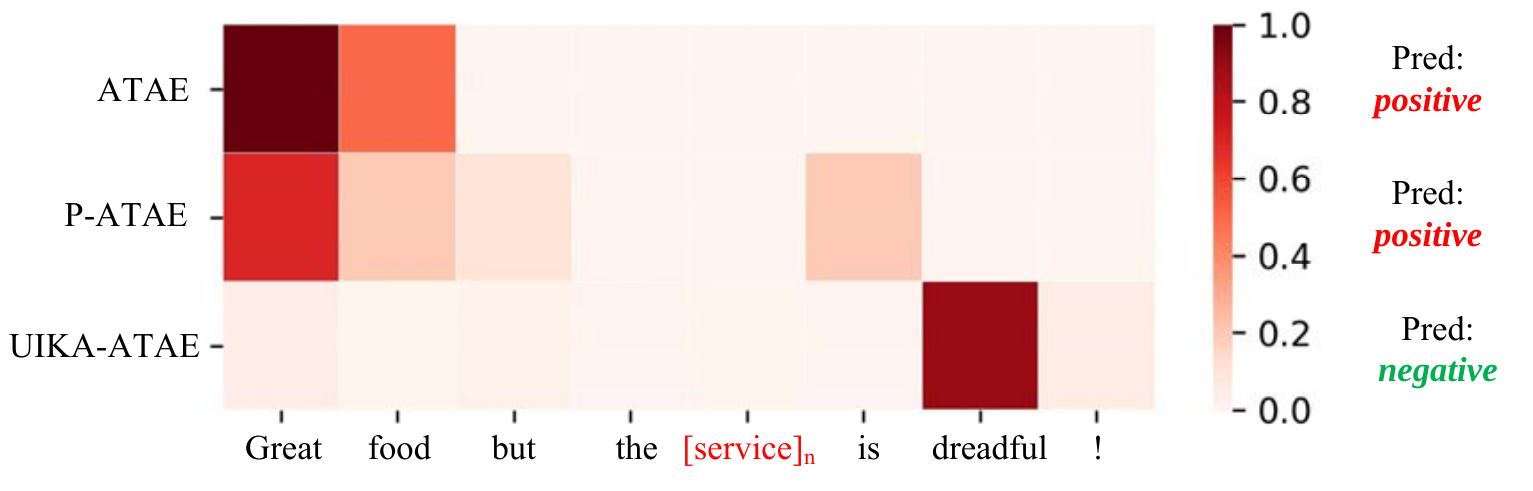} 
	\caption{The attention scores of ATAE-LSTM on a case from Restaurant14, where the aspect term is \emph{\textbf{service}} and the ground truth is \emph{\textbf{negative}}.}
	\label{fig2}
\end{figure}

\begin{figure}[tp]
	\centering
	\includegraphics[width=0.47\textwidth]{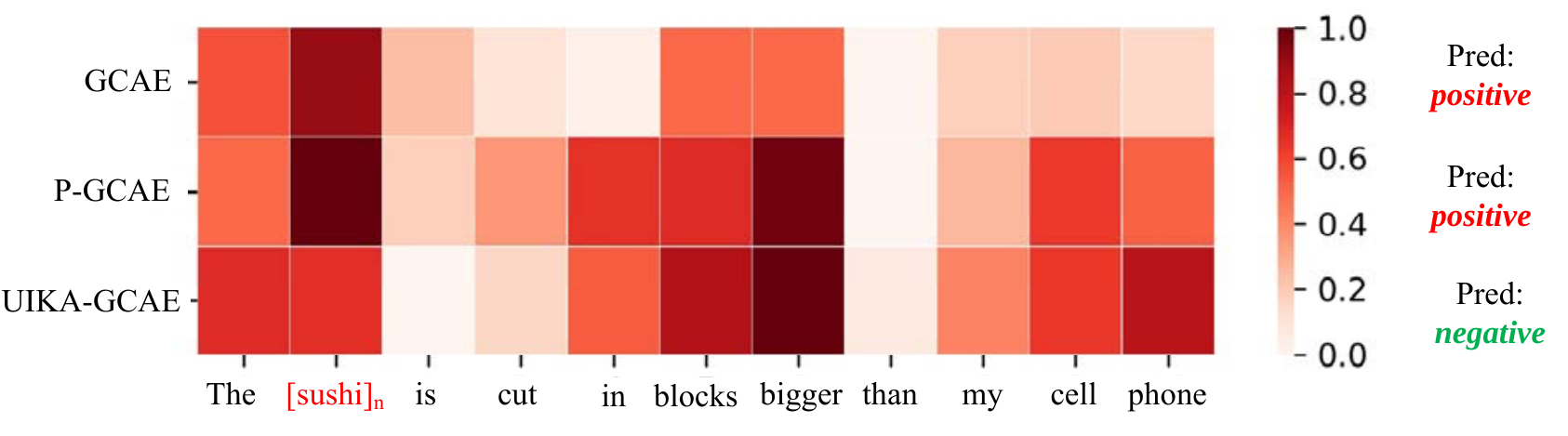} 
	\caption{Visualization of the gated outputs of GCAE. Notably, in this case, the aspect term is \emph{\textbf{sushi}} and the ground truth is \emph{\textbf{negative}}.}
	\label{fig3}
\end{figure}

Moreover, to exactly show how UIKA transfers the prior knowledge learned from the source datasets, we select two cases from the Restaurant14 dataset and employ ATAE-LSTM and GCAE as the baseline models to analyse them in detail. In practice, the cases are listed as follows:
	\begin{enumerate}
		\item Great food but the $[service]_n$ is dreadful!
		\item $[sushi]_n$ is cut into blocks larger than my cell phone.
	\end{enumerate}

First, for the first example, we visualize the attention scores of ATAE-LSTM towards aspect term \emph{\textbf{service}} and show the results in Fig.~\ref{fig2}. Specifically, the original ATAE-LSTM pays considerable attention to the unrelated opinion word ``Great", resulting in a false prediction.
In contrast, owing to pretraining on a large source dataset with rich sentence patterns, ATAE-UIKA has learned the characteristics of turning sentences and correctly selects the related opinion word ``dreadful" for \emph{\textbf{service}}. We also observe that P-ATAE can capture the important words ``but" and ``dreadful", but its main focus is still on ``Great", which leads to an incorrect prediction. This result validates that directly finetuning the pretraining model on the target dataset causes the prior knowledge to be forgotten.

Then, in the second case, we sum the outputs of the gated units in GCAE as the weight for each word and visualize the weights in Fig.~\ref{fig3}. Specifically, due to the implicit opinion words in the sentence, it is difficult for the models to directly capture the aspect-specific contextual features. Additionally, ``bigger" is a multiaffective word (\textit{i.e.} refer to \textit{positive} polarity in some cases but \textit{negative} polarity for the other cases), and there are only 15 instances containing ``big" in the training dataset. The above challenges prevent GCAE from learning the correct meaning of ``bigger", leading to an incorrect prediction. However, for GCAE-UIKA, since the pretraining Amazon dataset contains 2804 related instances that contain ``bigger" with various contexts and 1397 instances contain ``cell phone", during pretraining on the Amazon dataset, GCAE-UIKA can fully learn the corresponding affective polarities of ``bigger" for specific aspect terms, thus facilitating correct predictions \emph{positive} in this case. Similarly, the P-GCAE also focuses on ``bigger" but it does not make the correct prediction. Presumably, this is because P-GCAE pays more attention to ``sushi", which is mostly positive in the target training dataset, leading to a miscomprehension of this sentence.

\section{Discussion}
\label{section5}
\subsection{The Aspect Extraction} Method
\label{section 5.1}
As mentioned in Sec.~\ref{section3}, we convert the sentence-level datasets to pseudo aspect-level datasets via a simple aspect extraction (denoted as AE) approach. For better understanding, in this subsection, we describe this approach in detail. To take a closer look, we select some reviews from the Amazon and Yelp datasets and show them in Tab.~\ref{table7}. In practice, we first apply a natural language toolkit (NLTK) \footnote{https://www.nltk.org/} to split each review into sentences and produce the part-of-speech tag for each word. Then, we extract the nouns and noun phrases as candidate aspect terms. To reduce the noise, we remove the stopwords in the candidate terms.
It should also be noted that we may extract more than one aspect term in a sentence with different sentiment polarities, as shown in the third review in Tab.~\ref{table7}. For this case, we count the frequency of each candidate word and select the most frequent word as the final aspect term. Specifically, the final aspect terms in the third sentence of the Amazon and Yelp reviews are \emph{\textbf{battery life}} and \emph{\textbf{food}}, respectively. 

After the above processes, we can obtain the pseudo aspect terms and annotate them with the sentence-level sentiment labels. 
We state that such a simple approach may introduce some noise, \textit{e.g.} incorrect aspect extraction and false annotation. To analyse the influence of the noisy aspect terms, we conduct a contrastive experiment as follows: ``W/o AE'': we directly use the sampled sentiment-level instances as the pretraining corpus; ``Random AE'': we randomly extract a word from the sentence as the aspect term for each instance; ``Ours'': the processes used in our work.

\begin{table*}[tp]
	\centering
	\caption{Analysis of different aspect extraction (AE) approaches. ``Random AE'' denotes that a word selected randomly from the sentence is used as the aspect term. ``SA'' denotes the final ABSA performance. We use the GCAE as baseline here.}
	\label{tab_ac}
	\begin{tabular}{lcccccccc}
		\Xhline{0.8pt}
		\multirow{2}{*}{\textbf{\zqh{Approach}}} & \multicolumn{4}{c}{\textbf{\zqh{Laptop14}}}    & \multicolumn{4}{c}{\textbf{\zqh{Restaurant14}}}         \\ \cmidrule(lr){2-5}  \cmidrule(lr){6-9}  
		&\zqh{Precision (AE)}        & \zqh{Recall (AE) }   &Acc.(SA)         & F1.(SA)      &\zqh{Precision (AE) }       & \zqh{Recall (AE)}          & Acc.(SA)         & F1.(SA)               \\  \hline \hline
		\zqh{W/o AE}       & \zqh{--} & \zqh{-- } & $\downarrow$0.63\% & $\downarrow$1.14\% & \zqh{-- }   & \zqh{-- }   & $\downarrow$0.31\%    & $\downarrow$0.48\% \\
		\zqh{Random AE}   & \zqh{20.80\%} & \zqh{20.53\%}  & $\downarrow$0.56\%  & $\downarrow$1.04\% & \zqh{22.16\% }   & \zqh{21.52\%}        & $\downarrow$0.44\%    & $\downarrow$0.62\%          \\ 
		\zqh{Ours}         & \zqh{72.15\%} &\zqh{75.21\%}  &76.09\% &70.66\%  &\zqh{74.32\%} &\zqh{68.15\%}   &81.16\%   &71.56\%   \\ \Xhline{0.8pt}
	\end{tabular}
\end{table*}

To have a closer look, we first report the results of the aspect extraction. In practice, since the pretraining sentence-level datasets do not have aspect labels, we simply measure the performance of the aspect extraction methods on the Laptop14 and Restaurant14 datasets. The results are shown in Tab.~\ref{tab_ac}. We find that compared to ``Random AE'', our method can achieve much better performance, indicating its effectiveness.

Tab.~\ref{tab_ac} shows the final ABSA results. We observe that ``W/o AE'' performs worse than our method, confirming the importance of introducing aspect information during pretraining for ABSA. On the other hand, ``Random AE'' also achieves poor performance, as such a simple process introduces considerable noise and thus hinders the effectiveness of pretraining. In summary, we consider that a more accurate aspect extraction approach can boost the performance of our UIKA, and it is worthwhile to carefully explore sophisticated approaches to correctly extract the aspect terms, which we leave for future work.

\begin{table}[t]
		\caption{Comparative results of different pretraining datasets. Notably, the $\uparrow$ denotes the relative increase of performance compared to the corresponding baseline models. \textbf{A} and \textbf{Y} refer to pretraining the Amazon and Yelp datasets, respectively.}
	\label{table8}
	\centering
	\begin{tabular}{cccccc}
		\Xhline{0.8pt}
		\multirow{2}{*}{\textbf{}}	&\multirow{2}{*}{\textbf{Models}} & \multicolumn{2}{c}{\textbf{Laptop14}}    & \multicolumn{2}{c}{\textbf{Restaurant14}}          \\ \cline{3-6} 
		& & Acc.(\%)         & F1.(\%)          & Acc.(\%)         & F1.(\%)                 \\  \hline \hline
		\multirow{3}{*}{\textbf{}} 
		&GCAE               & 73.56          & 67.84          & 79.27          & 67.66     \\
		&ATAE               &70.66   &   64.95      & 76.44  &64.46       \\
		&ASGCN        & 75.55          & 71.05          & 80.77          & 72.02                   \\	\hline
		\multirow{3}{*}{\textbf{A}} 
		&GCAE-UIKA        & $\uparrow$2.56  & $\uparrow$2.82  & $\uparrow$1.89    & $\uparrow$3.90           \\ 
		&ATAE-UIKA       & \textbf{$\uparrow$5.28}   & \textbf{$\uparrow$5.19}     &\textbf{$\uparrow$4.46}     & \textbf{$\uparrow$5.39}         \\  
		&ASGCN-UIKA       & $\uparrow$3.82 & $\uparrow$5.00 & $\uparrow$3.61    & \textbf{$\uparrow$4.77}                \\	\hline
		\multirow{3}{*}{\textbf{Y}} 
		&GCAE-UIKA        &$\uparrow$1.75   &$\uparrow$2.06   &$\uparrow$1.98     & $\uparrow$3.53           \\ 
		&ATAE-UIKA       &$\uparrow$2.00  &$\uparrow$3.30   & $\uparrow$2.72    &  $\uparrow$3.82        \\  
		&ASGCN-UIKA       &$\uparrow$2.39   &$\uparrow$3.33  &$\uparrow$2.93    &$\uparrow$4.18              \\	\hline
		\multirow{3}{*}{\textbf{\zqh{A+Y}}} 
		&\zqh{GCAE-UIKA}       & \zqh{\textbf{$\uparrow$2.68}}  & \zqh{\textbf{$\uparrow$2.92}}  & \zqh{\textbf{$\uparrow$1.99}}    & \zqh{\textbf{$\uparrow$4.01}}           \\ 
		&\zqh{ATAE-UIKA}       & \zqh{$\uparrow$5.12}   & \zqh{$\uparrow$5.05}     &\zqh{$\uparrow$4.23}     & \zqh{$\uparrow$5.03}         \\  
		&\zqh{ASGCN-UIKA}       & \zqh{\textbf{$\uparrow$4.12}} & \zqh{\textbf{$\uparrow$5.11}} & \zqh{\textbf{$\uparrow$3.71}}    & \zqh{$\uparrow$4.53 }               \\	\hline
		\Xhline{0.8pt}
	\end{tabular}
\end{table}


\subsection{Influence of Pretraining Datasets}
\label{sec_pd}
Tab.~\ref{table1} shows the effectiveness of our proposed UIKA pretraining framework using the Amazon pretraining dataset. To further demonstrate that UIKA is not sensitive to the pretraining datasets, we also conduct extensive experiments on another pretraining dataset, \textit{i.e.} the Yelp dataset. Tab.~\ref{table8} reports the results pretrained on the Amazon and Yelp datasets.

We observe that not only do the models pretrained on the Amazon dataset achieve excellent improvements with UIKA but also the ones pretrained on the Yelp dataset, demonstrating that the UIKA framework is robust and not sensitive to the pretraining datasets. In particular, while pretraining on the Amazon dataset, our UIKA framework achieves a higher improvement on the Laptop14 dataset than on the Restaurant14 dataset. In contrast, while pretraining on the Yelp dataset, the performance improvement on Restaurant14 is higher than that on Laptop14. For instance, for the baseline ASGCN, the performance improvement of the \textbf{A} row on Laptop14 (\textit{Acc: +3.82\%, F1: +5.00\%}) is higher than that of Restaurant14 (\textit{Acc: +3.61\%, F1: +4.77\%}), while the improvement of the \textbf{Y} row on Laptop14 (\textit{Acc: +2.39\%, F1: +3.33\%}) is lower than that of Restaurant14 (\textit{Acc: +2.93\%, F1: +4.18\%}). One possible reason is that the Amazon pretraining dataset is more domain-similar to the Laptop14 dataset, and the Yelp is more domain-similar to the Restaurant14 dataset. These results validate the significance of bridging the domain gap between pretraining and the target ABSA datasets.

\zqh{Additionally, we also attempt to evaluate the effectiveness of UIKA on the combination of both pretraining datasets. The results are also listed in Tab.~\ref{table8}. It can be seen that more pretraining datasets basically produce more performance gains. However, since combining both pretraining datasets will introduce extra computational overhead, we use a single Amazon dataset as the pretraining dataset in our main experiments.
}

\section{Conclusion}
\label{section6}
In this paper, we propose a unified instance and knowledge alignment pretraining (UIKA) framework for ABSA. The proposed UIKA has three stages. The first stage presents a coarse-to-fine retrieval strategy to sample the pretraining instances and learns prior knowledge from the sampled instances. The second stage introduces a knowledge guidance learning strategy to better transfer prior knowledge and helps to learn domain-invariant knowledge, and the final stage finetunes the model obtained from the second stage to achieve a better performance on the target ABSA dataset. We conduct experiments using several different baseline methods on 
\zqh{six widely used ABSA}
datasets to demonstrate the effectiveness of our UIKA framework. The experiments show that UIKA outperforms the previous pretraining strategies by a wide margin, facilitating the baseline models to attain remarkable performances on most datasets. Encouragingly, we show that our UIKA functions as a plugin strategy, and consistently improves the existing competitive baseline methods. UIKA is a preliminary exploration of task-specific pretraining for the ABSA task, and it will be interesting to explore more effective learning strategies in the future.

\ifCLASSOPTIONcaptionsoff
  \newpage
\fi

\bibliographystyle{IEEEtran}
\bibliography{TASLP.bib}

\end{document}